\documentclass{article}

\PassOptionsToPackage{numbers}{natbib}
\usepackage[preprint]{neurips_2026}

\usepackage[utf8]{inputenc} 
\usepackage[T1]{fontenc}    
\usepackage{hyperref}       
\usepackage{url}            
\usepackage{booktabs}       
\usepackage{amsfonts}       
\usepackage{nicefrac}       
\usepackage{microtype}      
\usepackage{xcolor}         

\usepackage{graphicx}
\usepackage{subcaption}
\usepackage{makecell}
\usepackage{multirow}

\newcommand{\Sphere}{\mathbb{S}}
\newcommand{\Hyp}{\mathbb{H}}
\newcommand{\SO}{\mathrm{SO}}
\newcommand{\SE}{\mathrm{SE}}
\newcommand{\Exp}{\mathrm{Exp}}
\newcommand{\Log}{\mathrm{Log}}

\usepackage{amsmath}
\usepackage{amssymb}
\usepackage{mathtools}
\usepackage{amsthm}

\usepackage[capitalize,noabbrev]{cleveref}

\theoremstyle{plain}
\newtheorem{theorem}{Theorem}[section]

\theoremstyle{definition}

\theoremstyle{remark}

\title{IsoAct: Geometry-Aware Post-hoc Debiasing \\ via Isometric Actions}

\author{%
  Sumin Park\textsuperscript{1}, Taero Kim\textsuperscript{2,*}, Minhyeong Cho\textsuperscript{2}, \\ \textbf{Subeen Park\textsuperscript{3}, Jong Hwan Lee\textsuperscript{4}, Kyungwoo Song\textsuperscript{2,*}} \\
  \textsuperscript{1}University of Wisconsin–Madison, \textsuperscript{2}Yonsei University, \textsuperscript{3}SK Shieldus, \textsuperscript{4}Arist Inc.\\
  \small{\texttt{park749@wisc.edu}, \texttt{taero.kim@yonsei.ac.kr}, \texttt{mhcho0226@yonsei.ac.kr}} \\ \small {\texttt{sbpark22@skshieldus.com}, \texttt{jonghwan.lee@arist.io}, \texttt{kyungwoo.song@yonsei.ac.kr}}
}

\begin{document}

\maketitle

\let\thefootnote\relax\footnotetext{\textsuperscript{*} Corresponding Authors.}

\begin{abstract}
  Learning representations that are robust to nuisance factors is essential for reliable deployment, but remains challenging when the representations lie on non-Euclidean manifolds. Existing debiasing methods typically assume Euclidean representations or apply linear interventions, which can distort the geometry of structured latent spaces. We propose IsoAct, a post-hoc framework that accounts for geometry in both nuisance identification and representation editing. It identifies nuisance-predictive directions using Fisher discriminant analysis (FDA) in tangent coordinates and applies corrections through manifold-native actions that preserve manifold membership. Controlled and real-world experiments on hyperspherical, hyperbolic, and $\mathrm{SE}(3)$ representations evaluate nuisance reduction, utility preservation, and manifold validity jointly. IsoAct reduces nuisance recoverability while retaining competitive utility. Its clearest joint result is on KLIFS, combining reduced kinase-group recoverability, strong structural prediction, and valid rigid poses. Code: \url{https://github.com/MLAI-Yonsei/IsoAct}

\end{abstract}

\section{Introduction}
Representations robust to nuisance factors are important for reliable deployment, fairness, and domain robustness. Post-hoc debiasing and concept erasure methods address this problem by editing pretrained representations after feature extraction, avoiding the cost of retraining the pretrained encoder \citep{bolukbasi2016man,ravfogel2020null,ravfogel2022linear,belrose2023leace}. However, this setting becomes more challenging when representations lie on non-Euclidean manifolds rather than unconstrained Euclidean spaces. 
CLIP-style vision-language embeddings are commonly normalized onto hyperspheres~\citep{radford2021learning}, hyperbolic image-text models such as MERU place embeddings on Lorentz hyperbolic manifolds~\citep{desai2023hyperbolic}, and kinase-ligand structural datasets such as KLIFS contain ligand poses that can be represented as rigid motions in \(\mathrm{SE}(3)\)~\citep{kooistra2016klifs}.
For such representations, debiasing requires reducing nuisance information while preserving downstream utility and manifold validity, measured through residuals of the defining constraints (Table~\ref{tab:target_manifolds_icml}).

Existing post-hoc methods typically realize debiasing through Euclidean editing operations, including projection-based subspace removal \citep{bolukbasi2016man,ravfogel2020null}, linear concept erasure \citep{ravfogel2022linear,belrose2023leace}, feature intervention \citep{jung2024unified}, and ambient-space edits \citep{zhao2025bias,gerych2026wring}. These approaches are simple and effective baselines, and several preserve selected constraints in specific settings. A key limitation for manifold-valued representations is that the editing operation is not generally defined as a native transformation of the target manifold. As a result, an edit that is valid in Euclidean feature space does not generally preserve unit norm on the hypersphere, Lorentz validity in hyperbolic space, or the orthogonality and determinant constraints of a rotation matrix of \(\mathrm{SE}(3)\). Manifold-aware and geometry-inspired debiasing methods have therefore been proposed \citep{kumar2021identifying,zhong2026fairt2v,rathore2026causal}. Still, they are often tied to a particular geometry, generative setting, representation space, or training-time objective, leaving open a general post-hoc approach that respects geometry both when identifying nuisance structure and carrying out the resulting edit. 

Geometry matters at both stages of post-hoc debiasing. Local motion is described by tangent spaces, whereas an arbitrary ambient displacement need not respect the manifold's constraints~\citep{absil2008optimization}. This motivates identifying nuisance structure in coordinates that reflect local geometry and realizing the correction through a valid motion on the manifold.

Building on this principle, we propose IsoAct, a post-hoc framework for editing manifold-valued representations. IsoAct uses Fisher discriminant analysis (FDA) in geometry-specific tangent coordinates to identify nuisance-predictive directions. A neutral target constructed from low-confidence training examples supplies the correction reference. The resulting sample-dependent actions preserve manifold validity: angular updates on the hypersphere, angular and translation actions on the Lorentz hyperboloid, and rotation, translation, or joint updates on $\mathrm{SE}(3)$.

We evaluate IsoAct on controlled geometric benchmarks and real-world representations. Alongside KLIFS kinase-ligand poses, we examine protected attributes on CLIP and MERU embeddings of FairFace~\citep{karkkainen2019fairface} and use Flickr30K text-to-image retrieval~\citep{plummer2015flickr30k} to assess their utility. The clearest joint gains occur on KLIFS, where IsoAct combines reduced kinase-group recoverability, strong structural utility, and valid rigid poses. The experiments support evaluating nuisance reduction, utility, and manifold validity together, rather than relying on any one metric alone.

\section{Related Work}
\paragraph{Post-hoc debiasing and concept erasure.} 
Post-hoc debiasing and concept erasure have been extensively studied as practical approaches to removing protected or nuisance information from pretrained representations~\citep{bolukbasi2016man,meade2022empirical}. INLP~\citep{ravfogel2020null} removes linearly decodable protected-attribute information by iteratively training linear probes and projecting representations onto the intersection of their nullspaces. RLACE~\citep{ravfogel2022linear} and LEACE~\citep{belrose2023leace} further formulate linear concept erasure through a constrained adversarial game and a closed-form oblique projection, respectively. SFID~\citep{jung2024unified} mitigates bias through selective feature imputation over bias-relevant embedding coordinates, whereas SPD~\citep{zhao2025bias} models bias as a linear subspace and removes it via subspace projection with neutral mean reinjection. These methods operate in Euclidean feature spaces. IsoAct instead uses FDA in tangent coordinates and applies corrections through actions native to the target manifold.

\paragraph{Geometry-aware debiasing for manifold-valued representations.}
Euclidean projection can cause information loss by erasing valid associations~\citep{dev2021oscar} and may leave residual bias in the embedding geometry beyond the targeted bias direction~\citep{gonen2019lipstick}, motivating geometry-aware debiasing methods.
PGD~\citep{kumar2021identifying} mitigates gender bias in Poincar\'e word embeddings through per-word Riemannian optimization, while FAIRT2V~\citep{zhong2026fairt2v} uses anchor-based geodesic transformations on the hypersphere for training-free text-to-video debiasing.
PGD and FAIRT2V specialize to their respective geometries and applications.
More closely related, WRING~\citep{gerych2026wring} replaces projection with rotations in VLM embedding space.
IsoAct instead couples tangent-space nuisance identification with native actions in a common framework for hyperspherical, hyperbolic, and $\mathrm{SE}(3)$ representations.

\paragraph{Geometric fairness and training-time invariance.}
Fairness and debiasing have also been studied from the perspectives of representation geometry and invariance~\citep{he2020geometric}. CMF~\citep{rathore2026causal} enforces preservation of local Riemannian geometry across factual and counterfactual representations induced by sensitive-attribute interventions, through training-time regularization of the decoder Jacobian and Hessian. In contrast to CMF, IsoAct operates post hoc on fixed pretrained representations, complementing settings where retraining is costly or model internals are inaccessible.

\section{Preliminaries}
\label{sec:problem}

\subsection{Problem Setting}
\label{subsec:posthoc_problem}

We study post-hoc debiasing of pretrained representations on a known manifold $\mathcal M$. Let $\mathcal D=\{(x_i,s_i,y_i)\}_{i=1}^{N}$ contain $N$ samples, where $x_i\in\mathcal M$ is a representation, $s_i$ is a categorical nuisance label, and $y_i$ denotes task-relevant information. The nuisance label may encode a protected attribute, such as gender or race, or another undesired attribute. We seek a transformation $T:\mathcal M\to\mathcal M$ that makes $s_i$ less linearly predictable from $T(x_i)$ while preserving utility associated with $y_i$. The edit is applied after feature extraction, without retraining the pretrained encoder, and must retain membership in $\mathcal M$.

\subsection{Target Manifolds}

Table~\ref{tab:target_manifolds_icml} lists the three geometries and their membership constraints. The hypersphere $\mathbb S^{d-1}\subset\mathbb R^d$ contains unit vectors, with $d$ ambient coordinates and $d-1$ degrees of freedom, as used for normalized vision-language embeddings~\citep{radford2021learning}. A Lorentz-hyperbolic representation $x=(x_0,v)\in\mathbb H_c^d$ has $d$ spatial coordinates $v\in\mathbb R^d$ and a positive geometric time coordinate $x_0=\sqrt{c^{-1}+\|v\|_2^2}$. This time coordinate is part of the representation, not a temporal variable. The model has curvature $-c$, $c>0$, and is used by MERU~\citep{desai2023hyperbolic}. A rigid pose $x=(R,t)\in\mathrm{SE}(3)$ consists of a rotation matrix $R\in\mathrm{SO}(3)$ and a translation $t\in\mathbb R^3$. Its six degrees of freedom describe orientation and position, although storing $R$ and $t$ uses twelve entries~\citep{murray2017mathematical,hartley2013rotation,kooistra2016klifs}.

\begin{table*}[t!]
\caption{Target manifolds and the membership constraints that admissible edits must preserve. The Lorentz model has curvature $-c<0$, with $\langle x,x\rangle_L=-x_0^2+\sum_{j=1}^{d}x_j^2$ and time coordinate $x_0$.}
\label{tab:target_manifolds_icml}
\centering
\setlength{\tabcolsep}{4pt}
\renewcommand{\arraystretch}{0.90}
\begin{tabular}{@{}llll@{}}
\toprule
Manifold & Representation & \multicolumn{2}{c}{Constraints} \\
\midrule
Hypersphere
& $x\in\mathbb{S}^{d-1}$
& 1) $\|x\|_2=1$
& \\

Lorentz hyperboloid
& $x\in\mathbb{H}^{d}_{c}$
& 1) $\langle x,x\rangle_L=-1/c$
& 2) $x_0>0$ \\

$\mathrm{SE}(3)$
& $x=(R,t)\in\mathrm{SE}(3)$
& 1) $R^\top R=I_3$
& 2) $\det(R)=1$ \\
\bottomrule
\end{tabular}
\end{table*}

\subsection{Tangent Coordinates}
\label{subsec:tangent_coordinates}

At a reference point $r\in\mathcal M$, the tangent space describes admissible infinitesimal motion on the manifold~\citep{absil2008optimization}. Geometry-specific coordinate maps express representations in a common linear space: $\xi_i=\phi_r(x_i)\in\mathbb R^D$, where $D$ counts the coordinates used by the map and need not equal the intrinsic dimension. Such coordinates enable linear estimation while accounting for local geometric structure.

\section{IsoAct: Geometry-Aware Identification and Manifold-Native Actions}
\label{sec:method}

\begin{figure}[t!]
    \centering
    \includegraphics[width=\linewidth]{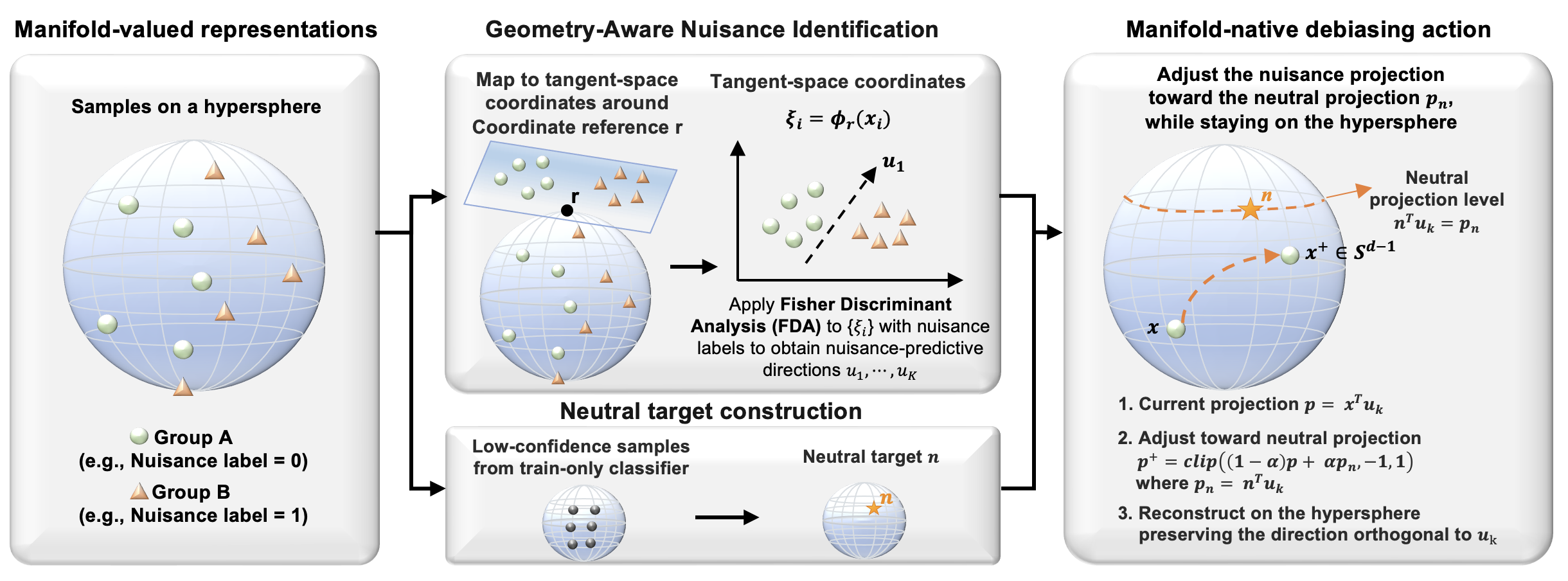}
    \caption{Overview of IsoAct on the hypersphere. FDA identifies nuisance directions in tangent coordinates around a coordinate reference $r$ constructed from training data. A neutral target $n$ from low-confidence training examples supplies the correction reference. The right panel shows one update along $u_k$ while preserving unit norm. Lorentz and $\mathrm{SE}(3)$ instances use the same pipeline with geometry-specific actions.}
    \label{fig:isoact_method}
\end{figure}

IsoAct identifies nuisance-predictive directions, constructs a neutral target, and applies corrections that remain on the manifold (Figure~\ref{fig:isoact_method}). The first two steps use training data; the resulting directions and target guide edits without retraining the representation encoder.

\subsection{Tangent-Space Nuisance Identification and Neutral Target}
\label{subsec:axes_reference_discrepancy}

\paragraph{Reference and coordinates.}
We first construct a coordinate reference $r\in\mathcal M$ from training representations: a normalized mean on the sphere, a mean spatial vector with its Lorentz time coordinate, or a mean pose with its rotation projected onto $\mathrm{SO}(3)$. This reference anchors the linearization, giving training coordinates $\xi_i=\phi_r(x_i)\in\mathbb R^D$. The coordinate maps use the manifold's logarithm, with transport or a basis change where required. Construction details are in Appendix~\ref{app:reference_neutral} through~\ref{app:se3_details_icml}.

\paragraph{Nuisance directions via FDA.}
Within these coordinates, $S_W$ measures spread among examples with the same nuisance label, whereas $S_B$ measures separation of their class means. FDA~\citep{fisher1936use,rao1948utilization} seeks directions along which this separation is large relative to within-class variation:
\begin{equation}
    \max_{u\in\mathbb{R}^D\setminus\{0\}}
    \frac{u^\top S_B u}
         {u^\top (S_W+\rho I_D)u},
    \label{eq:fda_objective}
\end{equation}
where $\rho>0$ regularizes the within-class scatter and $I_D$ is the $D$-dimensional identity. We retain $K$ directions $u_1,\ldots,u_K\in\mathbb R^D$ from the resulting discriminant subspace to describe nuisance-predictive variation. Appendix~\ref{app:axis_estimation_icml} gives the scatter and basis conventions.

\paragraph{Neutral target.}
The neutral target $n\in\mathcal M$ supplies the reference state for correction along the learned directions. An auxiliary nuisance classifier fitted on the training representations identifies examples whose nuisance labels it predicts with low confidence. We aggregate these examples using the corresponding manifold construction to form $n$ (Appendix~\ref{app:reference_neutral}). Thus $r$ anchors the analysis coordinates, while $n$ supplies the nuisance-related values used for correction. An action uses these components to compute its correction; it does not move every representation to $n$ itself.

\subsection{Geometry-Specific Action Primitives}
\label{subsec:geometry_action_primitives}

For each retained direction, IsoAct measures a nuisance-related discrepancy from $n$ and realizes the correction through a valid motion on the target manifold. Below, $x$ is the current representation and $x^+$ is the result of one action associated with $u_k$. The output becomes the input for the next direction; $r$, $n$, and the learned directions remain fixed. Nonnegative edit-strength hyperparameters control the correction magnitude and are selected by the validation rule in Section~\ref{subsec:selection_scope}.

\paragraph{Hypersphere.}
Changing a component of a unit vector generally changes its norm. IsoAct instead adjusts the nuisance projection and reconstructs the remaining component to stay on the sphere. Let $x,n\in\Sphere^{d-1}$ and represent the learned tangent direction as a unit vector $u_k\in\mathbb R^d$. For $|x^\top u_k|<1$, define the scalar projections of $x$ and $n$ onto $u_k$ and the normalized orthogonal component:
\[
    p\coloneqq x^\top u_k,\qquad p_n\coloneqq n^\top u_k,\qquad
    h\coloneqq\frac{x-pu_k}{\sqrt{1-p^2}}.
\]
With strength $\alpha\ge0$, the update is
\begin{equation}
    p^+=\operatorname{clip}\!\left((1-\alpha)p+\alpha p_n,-1,1\right),
    \qquad
    x^+=p^+u_k+\sqrt{1-(p^+)^2}\,h .
    \label{eq:sphere_action}
\end{equation}
At $\alpha=0$ the input is unchanged; $\alpha=1$ matches the neutral projection, and $\alpha>1$ permits extrapolation. Clipping keeps the requested projection feasible. Since $u_k$ and $h$ are orthonormal, reconstruction preserves unit norm while retaining the orthogonal direction $h$. This is a sample-dependent rotation in their span. A vanishing orthogonal component gives the identity action.

\paragraph{Lorentz hyperboloid.}
Here $x=(x_0,v)$ couples spatial coordinates $v$ to the geometric time coordinate $x_0$. Write $n=(n_0,n_{\mathrm{sp}})$ for the neutral target, and let $u_k\in\mathbb R^d$ be the unit spatial action direction supplied by FDA in the common tangent frame. IsoAct offers two motions. An \emph{angular} action adjusts $v^\top u_k$ relative to $n_{\mathrm{sp}}^\top u_k$ at fixed spatial radius $\|v\|_2$, using the spherical reconstruction above. Preserving this radius also preserves $x_0$ and the Lorentz constraint.

A \emph{hyperbolic translation} can change spatial radius, but must update $x_0$ and the axis projection together. With $p=v^\top u_k$ and strength $\alpha_{\mathrm{tr}}\ge0$, it requests $q_{\mathrm{tr}}=(1-\alpha_{\mathrm{tr}})p+\alpha_{\mathrm{tr}}n_{\mathrm{sp}}^\top u_k$ through a Lorentz boost:
\begin{equation}
    \begin{bmatrix}x_0^+\\p^+\end{bmatrix}
    =
    \begin{bmatrix}\cosh\tau&\sinh\tau\\\sinh\tau&\cosh\tau\end{bmatrix}
    \begin{bmatrix}x_0\\p\end{bmatrix},
    \qquad v^+=v+(p^+-p)u_k.
    \label{eq:hyperbolic_boost}
\end{equation}
The boost parameter $\tau$ controls this coupled motion, preserving $-x_0^2+p^2$ and the orthogonal spatial component. Appendix~\ref{app:lorentz_details} gives the target-to-$\tau$ calculation, the angular reconstruction domain, and numerical safeguards. The angular strength $\alpha_{\mathrm{ang}}\ge0$ and translation strength are independent; when combined, the second primitive acts on the first's output within each axis.

\paragraph{\texorpdfstring{$\boldsymbol{\SE(3)}$}{SE(3)}.}
For a pose $x=(R,t)$, translation may move in $\mathbb R^3$, whereas rotation must remain in $\SO(3)$. IsoAct therefore adds a translation increment to $t$ and multiplies $R$ by a valid rotation increment. The mapping in Appendix~\ref{app:se3_details_icml} converts each FDA direction $u_k\in\mathbb R^6$ into a unit pose direction $\bar u_k\in\mathbb R^{12}$ and native directions $a_k,d_k\in\mathbb R^3$ for rotation and translation. Nonzero native directions are normalized; a zero component disables that branch.

Let $n=(R_n,t_n)$ and let $\operatorname{pack}(x)\in\mathbb R^{12}$ concatenate the row-wise entries of $R$ with $t$. Translation is driven by the full pose discrepancy along $\bar u_k$:
\begin{equation}
    \delta^T=\left\langle\operatorname{pack}(n)-\operatorname{pack}(x),\bar u_k\right\rangle,
    \qquad t^+=t+\alpha\delta^T d_k.
    \label{eq:se3_translation_update}
\end{equation}
For rotation, $R^\top R_n$ compares the current and neutral orientations. Its rotation logarithm, converted to an axis-angle vector by $\vee$, yields the correction component along $a_k$:
\begin{equation}
    \delta^R=\left\langle\bigl(\Log_{\SO(3)}(R^\top R_n)\bigr)^\vee,a_k\right\rangle,
    \qquad R^+=R\Exp_{\SO(3)}\!\left(\alpha\delta^R\widehat a_k\right).
    \label{eq:se3_rotation_update}
\end{equation}
The hat map forms a skew-symmetric matrix from a vector, and $\vee$ is its inverse. Exponentiating this matrix produces a proper rotation, so multiplication keeps $R^+\in\SO(3)$. Rotation-only or translation-only edits keep the other component fixed; a joint step computes both discrepancies from the same incoming pose and returns $(R^+,t^+)$.

\begin{theorem}[Pointwise manifold validity]
\label{thm:pointwise_validity}
In exact arithmetic, each well-defined IsoAct update above with finite action parameters maps a valid input to the same target manifold. Any finite sequence of such updates preserves manifold membership, provided every step is defined.
\end{theorem}
\begin{proof}
The hyperspherical construction preserves unit norm; Lorentz spatial rotations and boosts preserve the quadratic constraint and upper sheet; and the $\SE(3)$ updates preserve $R\in\SO(3)$ and $t\in\mathbb{R}^3$. Applying these properties at each step gives the result. A detailed derivation of the $\SE(3)$ case is provided in Appendix~\ref{app:se3_update_icml}.
\end{proof}

\paragraph{Scope of the guarantee.}
Theorem~\ref{thm:pointwise_validity} guarantees membership of each edited point, not preservation of pairwise distances: correction magnitudes depend on the individual input.

\subsection{Validation-Based Configuration Selection}
\label{subsec:selection_scope}

For each real-world setting, we select one configuration per method using a common validation rule. Let $\mathcal{P}$ be the nuisance-probe metric set and $V_m(\theta)$ the validation value of metric $m$ for candidate $\theta$. Let $O_m$ and $N_m$ be the corresponding validation values for the unedited Original and a shuffled-label Null reference. Null provides the target level of nuisance predictability under label randomization. We summarize nuisance reduction by
\[
    P(\theta)
    =\frac{1}{|\mathcal{P}|}\sum_{m\in\mathcal{P}}
    \left(
        \lvert O_m-N_m\rvert
        -\lvert V_m(\theta)-N_m\rvert
    \right),
\]
so that larger values indicate closer agreement with Null relative to Original, including a penalty for overshooting Null. Let $U(\theta)$ denote the aggregate validation utility. We select the configuration maximizing $S(\theta)=[P(\theta)+U(\theta)]/2$ and use it for test reporting. The experiment-specific metrics and utility aggregation are specified in Appendix~\ref{app:selection_details}.

\section{Experiments}
\label{sec:experiments}

\begin{figure*}[t]
    \centering
    \includegraphics[width=0.88\textwidth]{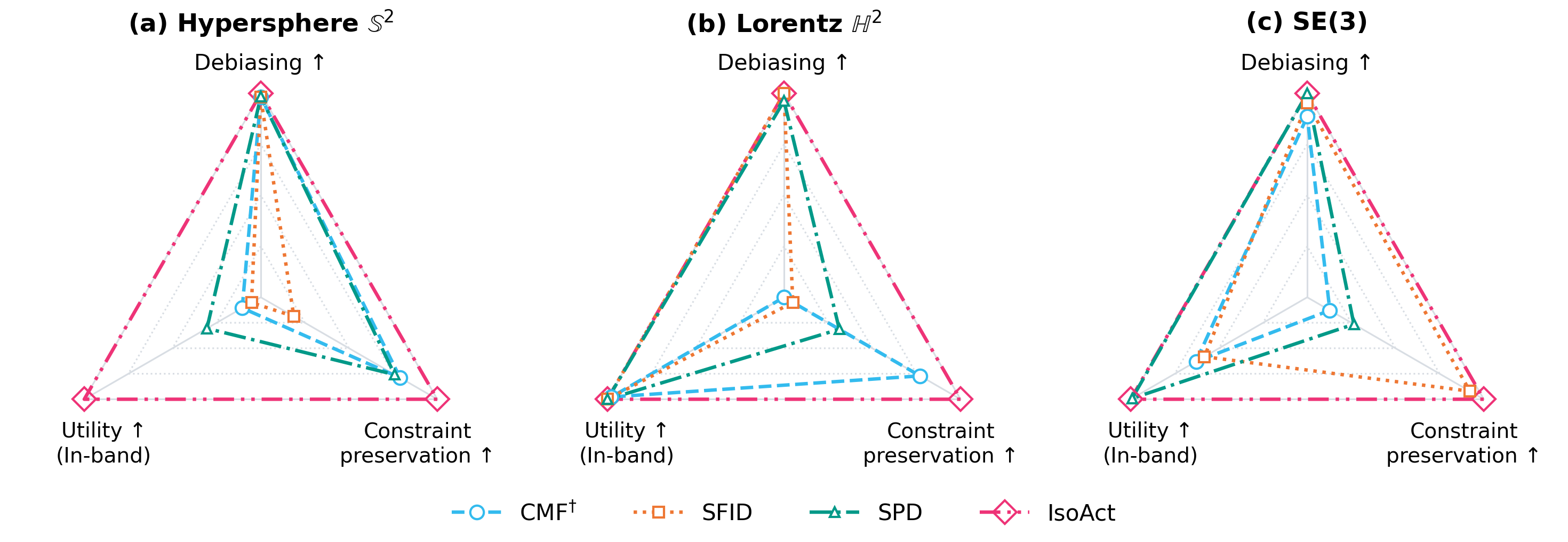}
    \caption{Controlled debiasing on three manifolds. IsoAct combines Oracle-matched nuisance-probe scores with preservation of the designated utility proxy and near-zero constraint residuals. All axes improve outward; constraint scales are panel-specific. $\dagger$ marks the learning-based CMF adaptation.}
    \label{fig:toy_starplots}
\end{figure*}

We evaluate post-hoc debiasing along three axes: \emph{nuisance reduction}, measured by linear-probe Accuracy and macro-F1; \emph{utility preservation}, measured on the designated tasks; and \emph{manifold validity}, measured by geometry-specific constraint residuals. We compare IsoAct with feature-intervention SFID~\citep{jung2024unified}, subspace-projection SPD~\citep{zhao2025bias}, and a CMF adaptation that learns an encoder-decoder with geometric regularization~\citep{rathore2026causal}.

The controlled benchmarks examine editing behavior under known nuisance and utility coordinates. We then evaluate real-world representations, beginning with KLIFS ligand poses and followed by CLIP and MERU embeddings. In the real-world benchmarks, transformations and evaluation predictors are fitted on training data, configurations are selected by the common validation rule in Section~\ref{subsec:selection_scope}, and selected results are reported on test data. Predictors are refitted to each transformed training set; retrieval instead uses fixed text embeddings.

\subsection{Controlled synthetic benchmarks}
\label{subsec:toy_experiments}

\subsubsection{Setup}
\label{subsubsec:toy_setup}

We construct mechanism checks on the hypersphere $\mathbb{S}^{2}$, the Lorentz hyperboloid $\mathbb{H}^{2}$ with curvature $-1$, and $\mathrm{SE}(3)$, with 400 samples split equally between two nuisance groups in each setting. These synthetic data have no semantic downstream task. Instead, a separate geometric variable, generated over the same range for both nuisance groups, serves as a utility proxy. Nuisance and utility occupy separate angular coordinates on the sphere; on Lorentz they occupy angular and radial coordinates. On $\mathrm{SE}(3)$, nuisance is injected into the rotation angle $\varphi$, while translation magnitude $\|t\|$ is the shared utility proxy. The goal is to remove group separation without disrupting this shared structure. Original and the known neutralized Oracle provide evaluation references, not competing methods.

Figure~\ref{fig:toy_starplots} reports normalized proximity of linear-probe Accuracy and macro-F1 to the Oracle, with Original at zero and Oracle at one. Probes use separate training and evaluation subsets. Utility measures the fraction satisfying a preservation criterion: a small polar-angle change on the sphere, or remaining within the original group-shared radial or translation-magnitude range on Lorentz and $\mathrm{SE}(3)$. Constraint preservation linearly rescales and reverses the mean residual, so smaller violations appear farther outward. Exact criteria, evaluation splits, panel scales, and additional results are in Appendix~\ref{app:controlled_scatter_icml}.


\subsubsection{Results}
\label{subsubsec:toy_results}

The controlled experiments make the central trade-off visible: strong nuisance reduction alone does not imply a successful manifold-valued edit. On the sphere, all three comparison methods nearly match the Oracle probe scores, yet at most $30.5\%$ of their outputs satisfy the polar-angle preservation criterion, compared with $100\%$ for IsoAct. On Lorentz, SFID and SPD preserve the task band and reduce nuisance predictability but substantially violate the hyperboloid constraint. On $\mathrm{SE}(3)$, SPD matches the Oracle probe scores and nearly preserves the task band while violating rotation constraints; SFID has smaller residuals but loses more utility. Across these constructions, IsoAct matches both Oracle probe metrics, satisfies the specified utility condition for all samples, and keeps the measured residuals at numerical precision.

\begin{figure}[t]
        \centering
        \includegraphics[width=\linewidth]{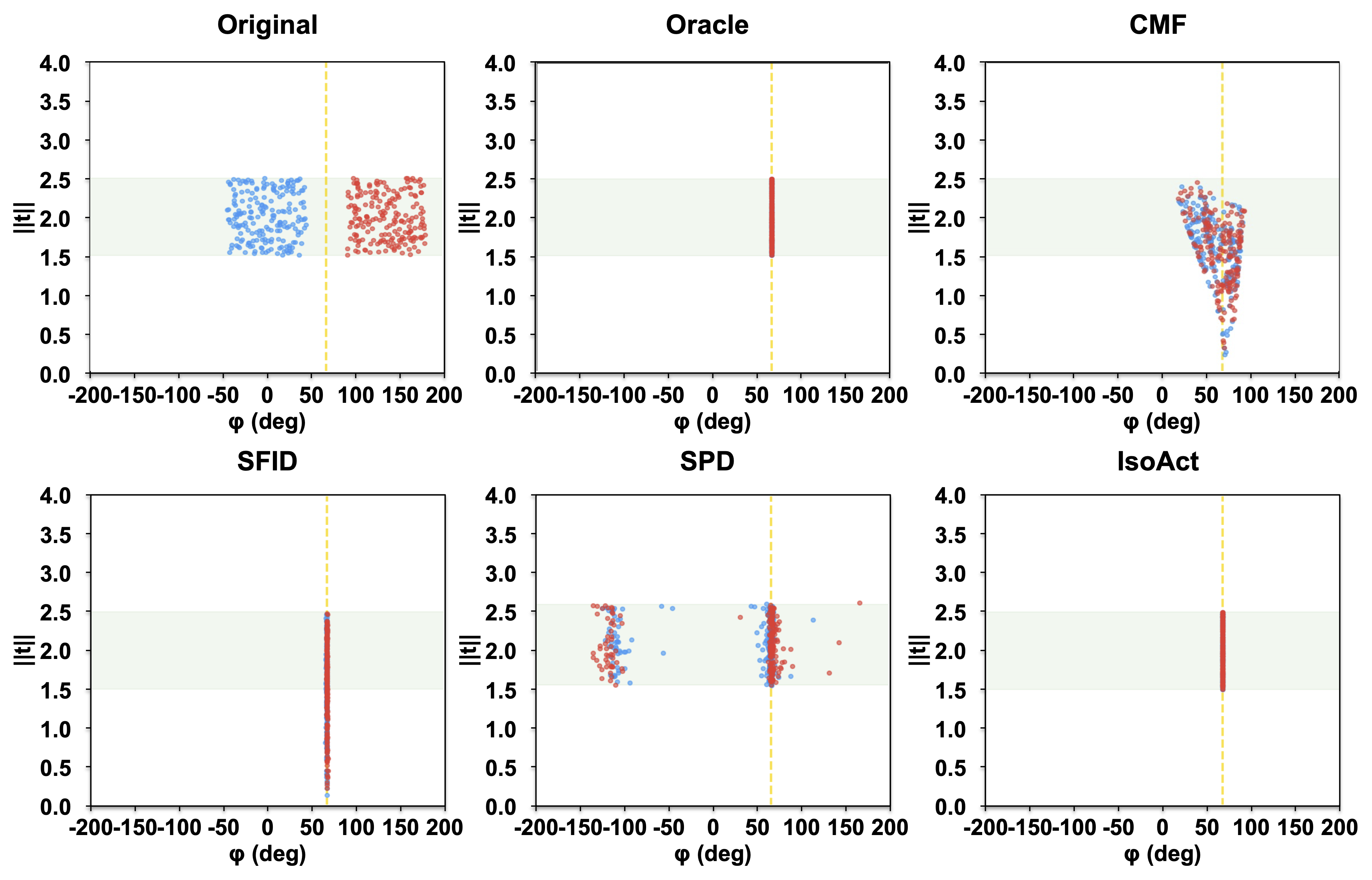}
        \caption{$\mathrm{SE}(3)$ mechanism check. Colors denote nuisance groups; the dashed line marks the neutral angle. The light-green band is the shared translation-magnitude range used as a utility proxy. IsoAct removes angular group separation while keeping all samples in this band, matching the Oracle pattern. In contrast, CMF and SFID move many samples outside the band. Rotation residuals and detailed results are given in Figure~\ref{fig:toy_starplots} and Appendix~\ref{app:controlled_scatter_icml}.}
        \label{fig:se3_scatter}
\end{figure}

Figure~\ref{fig:se3_scatter} provides a complementary point-level view: IsoAct collapses the injected group separation along the nuisance rotation coordinate while retaining the task-relevant translation band. The coordinate view and starplots together show why both task preservation and manifold residuals must accompany nuisance-probe evaluation. These are mechanism checks under controlled geometric structure, motivating the real-world tests that follow.

\subsection{Real-world \texorpdfstring{$\mathrm{SE}(3)$}{SE(3)} ligand-pose experiment on KLIFS}
\label{subsec:klifs_experiment}

\paragraph{Setup.}
We first evaluate IsoAct on KLIFS kinase-ligand structural data~\citep{kooistra2016klifs}. Each example is a ligand pose $x=(R,t)\in\mathrm{SE}(3)$, with orientation $R$ and position $t$ in the binding site~\citep{shen2021impact,zhang2022e3bind}. The nuisance label is one of nine kinase groups, a coarse grouping related to kinase-family identity~\citep{modi2019structurally}. We use it as a nuisance proxy to discourage reliance on family-specific cues while retaining structural information about conformation and ligand placement. Kinase IDs are disjoint across the training, validation, and test splits.

Utility comprises $\alpha$C-helix state classification and pocket-occupancy classification~\citep{kooistra2016klifs}. Linear probes measure kinase-group recoverability; separate classifiers fitted on the edited representations evaluate the structural tasks. We report nuisance Accuracy and macro-F1, utility Recall and AUROC, and mean rotation orthogonality and determinant residuals. Recall targets the training-set minority class; Pocket metrics average the front, gate, and back tasks. Original denotes unedited poses, and Null supplies the shuffled-label baseline targeted by nuisance reduction. Evaluation details and selected settings are given in Appendix~\ref{app:klifs_full_results_icml}.

\paragraph{Results.}

\begin{table*}[!t]
    \centering
    \caption{KLIFS results. IsoAct combines reduced kinase-group predictability with strong structural utility and numerical-precision rotation residuals. Constraints 1 and 2 report mean rotation orthogonality and determinant errors, respectively. Bold/underline mark best/second-best values among debiasing methods; Original and Null are reference rows excluded from ranking.}
    \label{tab:real_world_full_icml}
    \small
    \setlength{\tabcolsep}{2pt}
    \renewcommand{\arraystretch}{0.95}
    \begin{tabular}{clcccccccc}
    \toprule
    & & \multicolumn{2}{c}{Debiasing Capability}
    & \multicolumn{4}{c}{Utility Preservation}
    & \multicolumn{2}{c}{\makecell{Manifold Constraint\\Violation}} \\
    \cmidrule(lr){3-4}
    \cmidrule(lr){5-8}
    \cmidrule(lr){9-10}
    & Method
    & \multicolumn{2}{c}{Kinase Group}
    & \multicolumn{2}{c}{$\alpha$C-helix State}
    & \multicolumn{2}{c}{Pocket Occupancy}
    & \multicolumn{1}{c}{Constraint 1}
    & \multicolumn{1}{c}{Constraint 2} \\
    \cmidrule(lr){3-4}
    \cmidrule(lr){5-6}
    \cmidrule(lr){7-8}
    \cmidrule(lr){9-10}
    &
    & Acc$\downarrow$ & F1$\downarrow$
    & Recall$\uparrow$ & AUROC$\uparrow$
    & Recall$\uparrow$ & AUROC$\uparrow$
    & Mean$\downarrow$ & Mean$\downarrow$ \\
    \midrule
    \multirow{6}{*}{\rotatebox[origin=c]{90}{KLIFS}}
    & \textcolor{gray}{Original}
    & \textcolor{gray}{0.1398}
    & \textcolor{gray}{0.1409}
    & \textcolor{gray}{0.3362}
    & \textcolor{gray}{0.7163}
    & \textcolor{gray}{0.7357}
    & \textcolor{gray}{0.9708}
    & \textcolor{gray}{6.5065e-08}
    & \textcolor{gray}{2.2077e-08} \\

    & \textcolor{gray}{Null}
    & \textcolor{gray}{0.0827}
    & \textcolor{gray}{0.0679}
    & \textcolor{gray}{---}
    & \textcolor{gray}{---}
    & \textcolor{gray}{---}
    & \textcolor{gray}{---}
    & \textcolor{gray}{---}
    & \textcolor{gray}{---} \\

    \cmidrule{2-10}
    & CMF
    & 0.1988
    & 0.1858
    & 0.3046
    & 0.7087
    & \underline{0.8173}
    & \textbf{0.9732}
    & 3.6900e-01
    & 1.1368e-01 \\

    & SFID
    & \underline{0.1398}
    & \underline{0.1409}
    & \underline{0.3247}
    & \underline{0.7176}
    & 0.7388
    & 0.9712
    & \underline{6.5065e-08}
    & \underline{2.2077e-08} \\

    & SPD
    & 0.1425
    & 0.1415
    & 0.2414
    & 0.7061
    & 0.7797
    & \underline{0.9729}
    & 9.4948e-01
    & 3.7450e-01 \\

    & IsoAct
    & \textbf{0.1117}
    & \textbf{0.1022}
    & \textbf{0.3822}
    & \textbf{0.7333}
    & \textbf{0.8174}
    & 0.9664
    & \textbf{6.9510e-16}
    & \textbf{3.5723e-16} \\
    \bottomrule
    \end{tabular}
\end{table*}

The common validation rule balances nuisance reduction with utility, rather than selecting for nuisance reduction alone. At the selected operating points in Table~\ref{tab:real_world_full_icml}, IsoAct is the only compared method to reduce both kinase-group probe metrics relative to Original, while improving three of the four reported structural-utility metrics. SFID leaves nuisance recoverability essentially unchanged, whereas CMF and SPD incur appreciable rotation-constraint violations without reducing either nuisance metric. IsoAct combines these gains with numerical-precision rotation residuals, consistent with its manifold-native action design. The small decrease in pocket AUROC is its remaining utility trade-off. Together, these results highlight the value of reducing nuisance predictability through geometry-valid actions while retaining useful structural information.

\begin{figure}[t]
    \centering
    \begin{minipage}[t]{0.40\linewidth}
        \vspace{0pt}
        \centering
        \includegraphics[width=0.85\linewidth]{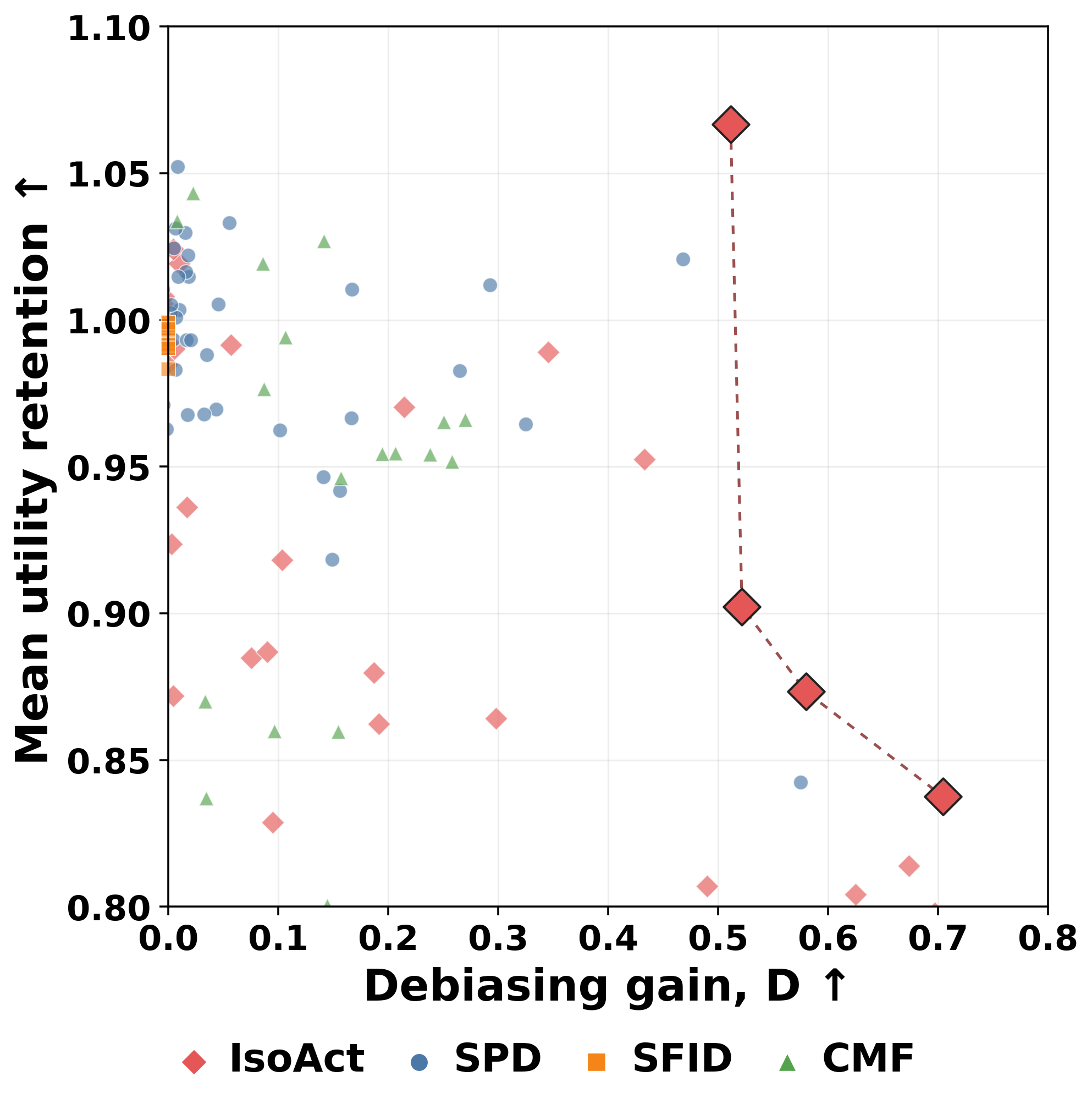}
    \end{minipage}
    \hfill
    \begin{minipage}[t]{0.59\linewidth}
    \centering
    \vspace{0pt}
        \caption{KLIFS debiasing-utility trade-offs. Each point is one configuration's test result; higher and farther right is better. Debiasing gain $D$ measures progress from Original ($0$) toward the shuffled-label Null ($1$). Mean utility retention quantifies utility preservation by averaging Recall and AUROC ratios to Original across both structural tasks; $1$ denotes unchanged aggregate utility. Outlined diamonds mark the IsoAct Pareto frontier. Its high-utility point combines $D\approx0.51$ with mean retention of $1.07$, while other frontier points achieve stronger debiasing at lower retention. Definitions and configuration pools are detailed in Appendix~\ref{app:pareto_details}.}
        \label{fig:se3_pareto}
    \end{minipage}
\end{figure}

Figure~\ref{fig:se3_pareto} extends the single-configuration comparison in Table~\ref{tab:real_world_full_icml}. For each nuisance metric $m\in\mathcal P=\{\mathrm{Acc},\mathrm{macro\mbox{-}F1}\}$, let $M_m$, $O_m$, and $N_m$ be the configuration, Original, and Null scores. Debiasing gain is $D=1-\frac12\sum_{m\in\mathcal P}|M_m-N_m|/|O_m-N_m|$. The method-specific pools cover overlapping edit-control ranges where applicable. IsoAct's frontier extends to approximately $D=0.70$, showing additional operating points beyond the table's selected setting and making the utility cost of stronger debiasing explicit.

\subsection{Real-world vision-language representations: CLIP and MERU}
\label{subsec:vlm_real_world}

\paragraph{Setup.}
We next evaluate the framework on unit-normalized CLIP ViT-B/32 image embeddings on the hypersphere~\citep{radford2021learning} and MERU ViT-B/16 image embeddings on a Lorentz hyperboloid~\citep{desai2023hyperbolic}. Each FairFace~\citep{karkkainen2019fairface} attribute, age, gender, or race, defines a separate experiment with fixed subsets of 3,500 training, 1,500 validation, and 1,500 test examples, identical across the two models. Linear probes measure protected-attribute recoverability. The FairFace-fitted transformation is also applied to Flickr30K image embeddings to evaluate text-to-image retrieval~\citep{plummer2015flickr30k}, with text embeddings fixed. Test retrieval uses 5,000 text queries and a gallery of 1,000 images.

In Table~\ref{tab:clip_meru_main}, Debias averages the normalized proximity of Accuracy and macro-F1 to the label-permutation Null reference using the Original-to-Null gap, so Original is zero and Null is one. Negative scores indicate greater average normalized distance from Null than Original. Utility averages Original-normalized R@1, R@5, and R@10, where R@$k$ measures how often a text query retrieves its target image among the top $k$ results. Protocols, normalization formulas, and selected settings are in Appendix~\ref{app:face_details}. Constraint residuals are measured on the FairFace test outputs, separately from retrieval: unit-norm error for CLIP and the mean of Lorentz quadratic and upper-sheet residuals for MERU (Appendix~\ref{app:constraint_diagnostics}).

\paragraph{Results.}

\begin{table*}[t]
    \centering
    \caption{CLIP and MERU results across protected attributes. IsoAct combines high retrieval retention with the strongest MERU debiasing and attribute-dependent CLIP gains. Debias averages normalized Accuracy/macro-F1 proximity to Null; Utility averages Original-normalized R@1/5/10. Constraint reports CLIP unit-norm or aggregated MERU Lorentz residuals. Bold/underline mark best/second-best method values, excluding reference rows and numerical-precision MERU residuals.}
    \label{tab:clip_meru_main}
    \setlength{\tabcolsep}{3.2pt}
    \renewcommand{\arraystretch}{1.05}
    \resizebox{\textwidth}{!}{%
    \begin{tabular}{llccccccccc}
    \toprule
    & & \multicolumn{3}{c}{Age} & \multicolumn{3}{c}{Gender} & \multicolumn{3}{c}{Race} \\
    \cmidrule(lr){3-5}\cmidrule(lr){6-8}\cmidrule(lr){9-11}
    Model & Method
    & Debias$\uparrow$ & Utility$\uparrow$ & Constraint$\downarrow$
    & Debias$\uparrow$ & Utility$\uparrow$ & Constraint$\downarrow$
    & Debias$\uparrow$ & Utility$\uparrow$ & Constraint$\downarrow$ \\
    \midrule

    \multirow{6}{*}{CLIP}
    & \textcolor{gray}{Original}
    & \textcolor{gray}{0.0000} & \textcolor{gray}{1.0000} & \textcolor{gray}{2.2612e-08}
    & \textcolor{gray}{0.0000} & \textcolor{gray}{1.0000} & \textcolor{gray}{2.2935e-08}
    & \textcolor{gray}{0.0000} & \textcolor{gray}{1.0000} & \textcolor{gray}{2.2181e-08} \\

    & \textcolor{gray}{Null}
    & \textcolor{gray}{1.0000} & \textcolor{gray}{---} & \textcolor{gray}{---}
    & \textcolor{gray}{1.0000} & \textcolor{gray}{---} & \textcolor{gray}{---}
    & \textcolor{gray}{1.0000} & \textcolor{gray}{---} & \textcolor{gray}{---} \\

    & CMF
    & 0.6339 & 0.0096 & 8.8207e-02
    & \underline{0.4481} & 0.0130 & 7.0113e-02
    & 0.6890 & 0.0115 & 1.1123e-01 \\

    & SFID
    & 0.0117 & \textbf{1.0005} & \underline{3.8070e-03}
    & 0.0389 & 0.9788 & \underline{8.1287e-03}
    & -0.0085 & \textbf{0.9975} & \underline{8.3426e-03} \\

    & SPD
    & \underline{0.7536} & 0.9867 & 1.8327e-02
    & \textbf{0.8243} & \underline{0.9938} & 1.0768e-02
    & \textbf{0.9656} & 0.9891 & 5.3183e-02 \\

    \cmidrule{2-11}
    & IsoAct
    & \textbf{0.9234} & \underline{0.9890} & \textbf{6.3230e-08}
    & 0.4254 & \textbf{1.0003} & \textbf{3.5303e-08}
    & \underline{0.8022} & \underline{0.9932} & \textbf{6.1190e-08} \\

    \midrule

    \multirow{6}{*}{MERU}
    & \textcolor{gray}{Original}
    & \textcolor{gray}{0.0000} & \textcolor{gray}{1.0000} & \textcolor{gray}{3.9909e-16}
    & \textcolor{gray}{0.0000} & \textcolor{gray}{1.0000} & \textcolor{gray}{4.0619e-16}
    & \textcolor{gray}{0.0000} & \textcolor{gray}{1.0000} & \textcolor{gray}{3.9435e-16} \\

    & \textcolor{gray}{Null}
    & \textcolor{gray}{1.0000} & \textcolor{gray}{---} & \textcolor{gray}{---}
    & \textcolor{gray}{1.0000} & \textcolor{gray}{---} & \textcolor{gray}{---}
    & \textcolor{gray}{1.0000} & \textcolor{gray}{---} & \textcolor{gray}{---} \\

    & CMF
    & 0.1039 & 0.2661 & 3.8310e-16
    & 0.2495 & 0.2936 & 3.9139e-16
    & 0.8001 & 0.0204 & 4.0323e-16 \\

    & SFID
    & 0.0759 & \textbf{0.9822} & 3.9257e-16
    & 0.0001 & \textbf{0.9944} & 3.8902e-16
    & 0.0466 & \textbf{0.9623} & 3.8192e-16 \\

    & SPD
    & \underline{0.9108} & \underline{0.9672} & 3.9198e-16
    & \underline{0.5836} & \underline{0.9869} & 3.8192e-16
    & \underline{0.8547} & 0.9321 & 3.9494e-16 \\

    \cmidrule{2-11}
    & IsoAct
    & \textbf{0.9392} & 0.9502 & 3.8902e-16
    & \textbf{0.8891} & 0.9692 & 3.8192e-16
    & \textbf{0.9825} & \underline{0.9449} & 3.9257e-16 \\

    \bottomrule
    \end{tabular}%
    }

\end{table*}

On CLIP, IsoAct retains at least $98.9\%$ mean retrieval utility across all attributes while maintaining unit norm to numerical precision. It achieves the strongest age debiasing and the second-highest race debiasing. SPD yields stronger gender and race debiasing, but with mean unit-norm errors of order $10^{-2}$. IsoAct thus offers nuisance reduction with minimal retrieval loss and valid hyperspherical outputs, a favorable combination not captured by ranking debiasing scores alone.

On MERU, IsoAct achieves the highest debiasing score for all three attributes while retaining approximately $94.5\%$ to $96.9\%$ of Original retrieval utility. Relative to SPD, stronger age and gender debiasing comes with lower retrieval retention, while race improves on both scores. SFID retains more retrieval utility but achieves little debiasing. Because MERU's time coordinate is reconstructed from its stored spatial vector to satisfy the hyperboloid constraint, all methods have numerical-precision residuals; nuisance reduction and retrieval provide the informative comparison here.

\subsection{Empirical summary and scope}
\label{subsec:empirical_summary}

The experiments support evaluating manifold-valued debiasing jointly through nuisance predictability, utility, and validity. The controlled checks illustrate the editing mechanism. KLIFS provides the clearest real-world joint result on a structured space of rotations and translations, combining reduced nuisance predictability with strong structural utility and valid poses. CLIP and MERU extend the evaluation to pretrained embedding spaces, with representation- and attribute-dependent trade-offs. Together, these results support the geometry-aware identification and manifold-native action framework across the three studied geometries. Reduced linear recoverability and small constraint residuals do not establish complete information removal, guaranteed fairness, or preservation of all pairwise distances.

\section{Conclusion}
\label{sec:conclusion}

We introduced IsoAct, a post-hoc debiasing framework that identifies nuisance-predictive directions using FDA in geometry-appropriate tangent coordinates and applies corrections through manifold-native actions that preserve manifold membership.
Controlled experiments expose cases where strong nuisance reduction coincides with lost task utility or invalid representations.
On KLIFS, IsoAct combines reduced linear kinase-group recoverability with strong structural utility and valid $\mathrm{SE}(3)$ poses.
CLIP and MERU further demonstrate its applicability to pretrained embeddings, with high retrieval retention and debiasing trade-offs that vary by model and protected attribute.
Together, these findings support treating nuisance reduction, utility preservation, and manifold validity as joint requirements for post-hoc representation editing.
Future work can extend this framework to learned or approximate manifolds and richer nuisance models.

{
\small
\bibliography{main}
\bibliographystyle{unsrt}
}


\appendix

\clearpage
\section{Limitations}
\label{app:limit_broader_icml}

IsoAct assumes a known target manifold and geometry-specific coordinate maps and action primitives. Extending the framework to learned or approximate geometries requires corresponding constructions. FDA identifies class-separating linear structure in tangent coordinates; nuisance information may remain accessible to nonlinear predictors or under distribution shift. The neutral target provides an editing reference rather than a guarantee of statistical independence or fairness.

The validity guarantee concerns manifold membership. Sample-dependent actions need not preserve pairwise distances, and valid poses do not by themselves establish physical feasibility or binding affinity. Our experiments assess designated structural and retrieval tasks in controlled, KLIFS, and vision-language settings; the observed trade-offs depend on the representation and nuisance attribute.

\section{Additional Method Details}
\label{app:method_details_icml}

\subsection{Procedure and notation}
\label{app:full_procedure}

Let $\mathcal I_{\mathrm{tr}}$ index the $N_{\mathrm{tr}}$ training examples, with representations $x_i\in\mathcal M$ and nuisance labels $s_i$. The training-derived reference $r$ defines coordinates $\xi_i=\phi_r(x_i)\in\mathbb R^D$, and $n\in\mathcal M$ is the neutral target. FDA yields retained directions $u_1,\ldots,u_K$. For $\SE(3)$, $u_k\in\mathbb R^6$ is mapped to a packed-pose direction $\bar u_k\in\mathbb R^{12}$ before constructing the native actions.

Each real-world candidate uses training data to construct $r$, estimate FDA directions, and form $n$ from low-confidence examples. These quantities remain fixed while editing all splits. Directions act sequentially on the current representation; additional passes repeat the same direction sequence. Evaluation predictors are fitted on the transformed training data, and the validation rule in Section~\ref{app:selection_details} selects the configuration used for test reporting. Retrieval uses fixed text embeddings in place of a fitted utility predictor.

\subsection{FDA nuisance-direction estimation}
\label{app:axis_estimation_icml}

Let $\mathcal C$ be the nuisance-label set, $C=|\mathcal C|$, and $\mathcal I_c=\{i\in\mathcal I_{\mathrm{tr}}:s_i=c\}$, with $N_c=|\mathcal I_c|$. Define class means $\mu_c=N_c^{-1}\sum_{i\in\mathcal I_c}\xi_i$ and the overall mean $\mu=N_{\mathrm{tr}}^{-1}\sum_{i\in\mathcal I_{\mathrm{tr}}}\xi_i$. The unnormalized scatter matrices are
\begin{align}
 W_0&=\sum_{c\in\mathcal C}\sum_{i\in\mathcal I_c}(\xi_i-\mu_c)(\xi_i-\mu_c)^\top,\\
 B_0&=\sum_{c\in\mathcal C}N_c(\mu_c-\mu)(\mu_c-\mu)^\top.
\end{align}
Their normalized forms define the regularized Fisher criterion in Eq.~(\ref{eq:fda_objective}).

For CLIP and MERU, $S_W=W_0/(N_{\mathrm{tr}}-C)$ and $S_B=B_0/(C-1)$. With $A_\rho=S_W+\rho I_D$ and $\rho>0$, binary FDA solves $A_\rho w=\mu_1-\mu_0$ for class labels $0,1$. Multiclass FDA diagonalizes $A_\rho^{-1/2}S_BA_\rho^{-1/2}$ and maps the resulting eigenvectors back through $A_\rho^{-1/2}$. The ridge $\rho$ is on the scale of the normalized within-class covariance.

For each ridge value, singular value decomposition converts the discriminant basis fitted at the largest requested rank to an orthonormal row basis. The transposes of its first $K$ rows define $u_1,\ldots,u_K$, with deterministic signs and numerical rank. The effective rank is bounded by $C-1$ and the tangent dimension. Each nuisance class has at least two training examples.

For $\SE(3)$, the six-dimensional coordinates instead use $S_W=W_0/N_{\mathrm{tr}}$ and $S_B=B_0/N_{\mathrm{tr}}$. We solve
\[
 S_Bu=\lambda(S_W+10^{-6}I_6)u
\]
in descending eigenvalue order, with a deterministic basis within tied eigenspaces. Six directions are constructed before selecting the action prefix. Section~\ref{app:se3_details_icml} specifies their lift to pose directions.

\subsection{Coordinate references and neutral targets}
\label{app:reference_neutral}
\label{app:neutral_reference}

The reference $r$ defines the linearization for FDA, whereas $n$ supplies the nuisance state used to compute corrections. Both are obtained from training data through separate constructions.

For CLIP, $r$ is the normalized arithmetic mean of the training unit vectors. For MERU, the reference retains the mean spatial vector and restores the positive time coordinate to satisfy the Lorentz constraint. For $\SE(3)$, use $r=(R_r,t_r)$ with
\[
 R_r=\operatorname{Proj}_{\SO(3)}\!\left(\frac1{N_{\mathrm{tr}}}\sum_{i\in\mathcal I_{\mathrm{tr}}}R_i\right),
 \qquad t_r=\frac1{N_{\mathrm{tr}}}\sum_{i\in\mathcal I_{\mathrm{tr}}}t_i.
\]
The determinant-corrected polar projection of $M=L\Sigma V^\top$ is $L\operatorname{diag}(1,1,\det(LV^\top))V^\top$.

An auxiliary nuisance classifier fitted on training data supplies class probabilities $q_c(x)$. Its low-confidence subset is
\[
 \mathcal I_{\mathrm{low}}=\{i\in\mathcal I_{\mathrm{tr}}:\max_c q_c(x_i)<0.7\}.
\]
For CLIP, $n$ is the normalized mean of this subset. For MERU, it is the mean spatial vector with the corresponding positive time coordinate. If fewer than ten examples qualify, these two constructions use the full-training mean. The $\SE(3)$ target averages the low-confidence poses and applies the polar projection to the mean rotation.

The MERU and KLIFS auxiliary classifiers use standardized logistic regression with inverse regularization strength 1, at most 2,000 iterations, and unweighted classes. They are fitted to the original training representations to construct the neutral target. These auxiliary classifiers are distinct from the balanced nuisance probe used for KLIFS evaluation.

\subsection{Hyperspherical coordinates and actions}
\label{app:sphere_details}
\label{app:geometry_actions_icml}

For unit vectors $x,r$, let $b=r^\top x$ and $\vartheta=\arccos b$. Away from the antipodal singularity,
\[
 \operatorname{Log}_r(x)=\frac{\vartheta}{\sin\vartheta}(x-br),
\]
with the continuous limit at $x=r$. Inner products are clipped for numerical evaluation, and coordinates and learned directions are projected onto $T_r\Sphere^{d-1}$ before direction normalization and orthogonalization. CLIP represents these tangent vectors in 512 ambient coordinates, with tangent dimension 511.

For a current unit representation and unit action direction $u_k$, define the current projection $p=x^\top u_k$, the neutral projection $p_n=n^\top u_k$, and the orthogonal unit direction $h=(x-pu_k)/\sqrt{1-p^2}$ when $|p|<1$. Equation~(\ref{eq:sphere_action}) adjusts the projection and reconstructs the point along $h$. When clipping is inactive, $\alpha=1$ matches $p_n$, while $\alpha>1$ extrapolates beyond it.

A vanishing orthogonal component gives the identity action. Numerical stabilization clips projections just inside the unit interval and renormalizes after an axis update only when unit-norm drift exceeds $10^{-5}$. Additional passes repeat the fixed axis sequence.

\subsection{Lorentz-hyperbolic coordinates and actions}
\label{app:lorentz_details}

Use $\langle x,y\rangle_L=-x_0y_0+x_{\mathrm{sp}}^\top y_{\mathrm{sp}}$ on $\Hyp_c^d$, with curvature $-c$ for $c>0$. A stored spatial vector $v\in\mathbb R^d$ represents $x=(\sqrt{c^{-1}+\|v\|^2},v)$. Given $r$, let $\beta=-c\langle r,x\rangle_L$. The logarithmic tangent vector is
\[
 \ell_r(x)=\frac{\operatorname{arcosh}\beta}{\sqrt{\beta^2-1}}(x-\beta r),
\]
using the continuous limit at $x=r$, numerical clipping, and tangent projection. To compare all samples in a common tangent frame, transport $\ell_r(x)$ to $o=(1/\sqrt c,0,\ldots,0)$. For $\eta\in T_r\Hyp_c^d$,
\[
 \operatorname{PT}_{r\to o}(\eta)=\eta+
 \frac{\langle o,\eta\rangle_L}{c^{-1}-\langle r,o\rangle_L}(r+o).
\]
The FDA coordinate is $\xi_i=[\operatorname{PT}_{r\to o}(\ell_r(x_i))]_{\mathrm{sp}}$, where $[\cdot]_{\mathrm{sp}}$ retains the spatial components. A tangent vector at $o$ has zero time component, yielding 512 FDA coordinates for MERU.

For a unit action direction $u_k\in\mathbb R^d$ and neutral target $n=(n_0,n_{\mathrm{sp}})$, set $p=v^\top u_k$, $z=v-pu_k$, $q=n_{\mathrm{sp}}^\top u_k$, and $R_v=\|v\|$. When $\|z\|>0$, the angular action with strength $\alpha_{\mathrm{ang}}\ge0$ is
\[
 p_{\mathrm{ang}}=\operatorname{clip}((1-\alpha_{\mathrm{ang}})p+\alpha_{\mathrm{ang}}q,-R_v,R_v),
 \qquad
 v^+=p_{\mathrm{ang}}u_k+
 \sqrt{R_v^2-p_{\mathrm{ang}}^2}\,\frac{z}{\|z\|}.
\]
It preserves spatial radius and the time coordinate.

For translation strength $\alpha_{\mathrm{tr}}\ge0$, let $q_{\mathrm{tr}}=(1-\alpha_{\mathrm{tr}})p+\alpha_{\mathrm{tr}}q$. A Lorentz boost gives $p^+=x_0\sinh\tau+p\cosh\tau$. With $a_+=x_0+p$, $a_-=x_0-p$, and $w=e^\tau$, the equation for the requested projection is
\[
 a_+w^2-2q_{\mathrm{tr}}w-a_-=0,
 \qquad
 w=\frac{q_{\mathrm{tr}}+\sqrt{q_{\mathrm{tr}}^2+a_+a_-}}{a_+}.
\]
Valid Lorentz inputs have $a_+,a_->0$, so the displayed root is the positive one and $\tau=\log w$. Numerical evaluation bounds rapidity to $[-8,8]$ and uses $\tau=0$ if no finite positive solution is obtained. The boost in Eq.~(\ref{eq:hyperbolic_boost}) updates time and the axis projection while leaving $z$ unchanged.

The angular and translation strengths are independent. Their configured order is applied within each axis, with the second primitive receiving the first primitive's output.

\subsection{\texorpdfstring{$\SE(3)$}{SE(3)} coordinates, direction lift, and actions}
\label{app:se3_details_icml}

For $x=(R,t)$, use $\operatorname{pack}(x)=[\operatorname{vec}_{\mathrm{row}}(R);t]\in\mathbb R^{12}$ with row-major vectorization and native translation units. Relative to $r=(R_r,t_r)$, the right-body pose logarithm has six-component coordinates $\zeta_i=(\omega_i,v_i)\in\mathbb R^6$. Its rotation component is $\omega_i=(\Log_{\SO(3)}(R_r^\top R_i))^\vee$, and its translation component is $v_i=J_l(\omega_i)^{-1}R_r^\top(t_i-t_r)$.

The left Jacobian $J_l$ couples rotation and translation: a twist $(\omega,v)$ exponentiates to rotation $\Exp_{\SO(3)}(\widehat\omega)$ and translation $J_l(\omega)v$. With $\vartheta=\|\omega\|$,
\[
 J_l(\omega)=I_3+
 \frac{1-\cos\vartheta}{\vartheta^2}\widehat\omega+
 \frac{\vartheta-\sin\vartheta}{\vartheta^3}\widehat\omega^2.
\]
Near zero, $J_l$ and $J_l^{-1}$ use the expansions $I_3+\widehat\omega/2+\widehat\omega^2/6$ and $I_3-\widehat\omega/2+\widehat\omega^2/12$, respectively.

The matrix $J_r\in\mathbb R^{12\times6}$ maps infinitesimal right-body perturbations at $r$ to packed-pose coordinates. For standard basis vectors $e_j\in\mathbb R^3$, its rotation columns are $[\operatorname{vec}_{\mathrm{row}}(R_r\widehat e_j);0]$ and its translation columns are $[0;R_re_j]$. Let $B_r\in\mathbb R^{12\times6}$ be a deterministic orthonormal basis of $\operatorname{range}(J_r)$. The FDA coordinates and lifted unit directions are
\[
 \xi_i=B_r^\top J_r\zeta_i,
 \qquad
 \bar u_k=\frac{B_ru_k}{\|B_ru_k\|},
 \qquad u_k\in\mathbb R^6.
\]
Thus $\zeta_i$ is the pose-logarithm coordinate and $\xi_i$ is the coordinate used by FDA. The native rotation direction $a_k\in\mathbb R^3$ is the normalized rotation component of $J_r^+\bar u_k$, where $J_r^+$ is the pseudoinverse. The translation direction $d_k\in\mathbb R^3$ is the normalized translation block of $\bar u_k$. A zero corresponding component disables that branch.

The translation action uses the full packed-pose discrepancy in Eq.~(\ref{eq:se3_translation_update}); the rotation action uses the relative rotation to $R_n$ in Eq.~(\ref{eq:se3_rotation_update}). Rotation-only and translation-only actions keep the other component fixed. A joint action computes both corrections from the same incoming pose, while successive axis steps use the preceding output. Numerical rotation updates use the polar projection defined in Section~\ref{app:reference_neutral} for stabilization.

\subsection{Detailed proof of \texorpdfstring{$\SE(3)$}{SE(3)} validity}
\label{app:se3_update_icml}

For the $\SE(3)$ case of Theorem~\ref{thm:pointwise_validity}, fix an axis and a valid incoming pose $x=(R,t)$. With the axis fixed, write the discrepancies in Eqs.~(\ref{eq:se3_translation_update}) and~(\ref{eq:se3_rotation_update}) as functions $\delta^T(x)$ and $\delta^R(x)$. Let $\eta_R(x)=\alpha\delta^R(x)$ and $\eta_T(x)=\alpha\delta^T(x)$ be their finite correction magnitudes. With action directions $a,d\in\mathbb R^3$, the three maps are
\begin{align}
 \Phi_R(x)&=(RQ(x),t),\\
 \Phi_T(x)&=(R,t+\eta_T(x)d),\\
 \Phi_{RT}(x)&=(RQ(x),t+\eta_T(x)d),
 \qquad Q(x)=\exp(\eta_R(x)\widehat a).
\end{align}

\begin{proof}
For $a=(a_1,a_2,a_3)^\top$, the matrix
\[
 \widehat a=\begin{bmatrix}
 0&-a_3&a_2\\a_3&0&-a_1\\-a_2&a_1&0
 \end{bmatrix}
\]
is skew-symmetric. Set $A=\eta_R(x)\widehat a$. Then
\[
 Q(x)^\top Q(x)=\exp(A^\top)\exp(A)=\exp(-A)\exp(A)=I,
 \qquad \det Q(x)=\exp(\operatorname{tr}A)=1.
\]
Thus $Q(x)\in\SO(3)$. Since $R\in\SO(3)$,
\[
 (RQ)^\top(RQ)=Q^\top R^\top RQ=I,
 \qquad \det(RQ)=\det R\,\det Q=1.
\]
Also $t+\eta_T(x)d\in\mathbb R^3$. Consequently, the rotation-only, translation-only, and joint maps all return an element of $\SE(3)$. Disabled branches are identities and satisfy the same property. Induction over any finite sequence of these maps proves the result, even when correction magnitudes are recomputed from the current pose at each step.
\end{proof}

\subsection{Validation-based configuration selection}
\label{app:selection_details}
\label{app:implementation_conventions_icml}

For each real-world setting, let $\mathcal P=\{\mathrm{Acc},\mathrm{macro\mbox{-}F1}\}$ and let $V_m(\theta)$ denote candidate $\theta$'s validation nuisance metric. The corresponding Original and shuffled-label Null values are $O_m$ and $N_m$. Null supplies the target level of nuisance predictability under label randomization. The nuisance component is
\[
 P(\theta)=\frac12\sum_{m\in\mathcal P}
 \bigl(|O_m-N_m|-|V_m(\theta)-N_m|\bigr).
\]
For CLIP and MERU, $U(\theta)$ is the arithmetic mean of validation R@1, R@5, and R@10. For KLIFS, it is the harmonic mean $H(F1_{\alpha C},F1_{\mathrm{Pocket}})$ of the two tasks' validation F1 scores, where $H(a,b)=2ab/(a+b)$ and $H(0,0)=0$. We maximize $S(\theta)=[P(\theta)+U(\theta)]/2$ among candidates with finite scores, breaking ties by a fixed deterministic ordering.

This rule balances nuisance reduction and utility within each method's candidate pool. Sections~\ref{app:klifs_full_results_icml} and~\ref{app:face_details} give the selected settings; Section~\ref{app:pareto_details} describes the additional configuration sweep.

\section{Experimental Details and Additional Results}
\label{app:experiments}

\subsection{Controlled synthetic benchmarks}
\label{app:controlled_scatter_icml}

\paragraph{Construction and evaluation.}
Each benchmark has 400 examples, equally divided between two nuisance groups, with seed 42. Nuisance is injected into an angular coordinate; a separate geometric variable generated over the same range for both groups serves as a utility proxy. The proxy is polar angle on $\Sphere^2$, a radial coordinate on the curvature-$-1$ Lorentz hyperboloid $\Hyp^2$, and translation magnitude on $\SE(3)$. Original and the known neutralized Oracle provide the evaluation references.

A stratified split with seed 42 gives 280 probe-training and 120 probe-evaluation examples. Logistic-regression nuisance probes use at most 1,000 iterations on the sphere and 500 with the \texttt{lbfgs} solver on Lorentz and $\SE(3)$; the $\SE(3)$ probe has inverse regularization strength $0.1$. Probe Accuracy and macro-F1 use the 120 evaluation examples, while utility and constraint residuals are measured over all 400 edited outputs.

\paragraph{Debiasing coordinate.}
Let $M_m$, $O_m$, and $Q_m$ denote nuisance metric $m$ for an edited representation, Original, and Oracle, respectively. The starplot uses
\begin{equation}
 D_{\mathrm{toy}}=1-\frac12\sum_{m\in\{\mathrm{Acc},\mathrm{macro\mbox{-}F1}\}}
 \frac{|M_m-Q_m|}{|O_m-Q_m|}.
 \label{eq:app_toy_debias}
\end{equation}
Each metric is normalized by its own Original-to-Oracle gap before averaging. Original maps to zero and Oracle to one; a score of one matches the Oracle's two probe metrics.

\paragraph{Utility coordinate.}
The utility score $U$ is the fraction of outputs satisfying the designated proxy-preservation criterion. On the sphere, let $\vartheta(x)$ be the polar angle in degrees, computed as $(180/\pi)\arccos(\operatorname{clip}(x_3/\|x\|,-1,1))$ for nonzero $x$. For norms below $10^{-10}$, the denominator is set to one. The criterion is $|\vartheta(x_i^{\mathrm{edited}})-\vartheta(x_i^{\mathrm{original}})|\le2^\circ$.

For Lorentz outputs, define the extracted radial coordinate $\varrho(x)=\operatorname{arcosh}(\max(x_0,1))$ and require $1.5\le\varrho(x_i^{\mathrm{edited}})\le2.5$. For $\SE(3)$, require $1.5\le\|t_i^{\mathrm{edited}}\|\le2.5$. These intervals are the utility-proxy ranges shared by the two groups. Thus the starplot summarizes original-sample tolerance on the sphere and interval retention in the other two settings.

\paragraph{Constraint coordinate.}
Let $\overline C_j$ be the sample mean of residual $C_j$, defined in Section~\ref{app:constraint_diagnostics}. Set $E=\overline C_1$ on the sphere and $E=(\overline C_1+\overline C_2)/2$ on Lorentz and $\SE(3)$. The displayed radius is $r_C=1-E/B$, with panel scales $B=0.10$, $8.00$, and $1.20$, respectively. Larger radii correspond to smaller violations. The scales support within-panel comparisons rather than absolute residual comparisons across manifolds.

\begin{table}[tbp]
\centering
\caption{Detailed results underlying the toy starplots. Acc and macro-F1 are ranked by proximity to Oracle; higher $D_{\mathrm{toy}}$ and $U$ and lower residuals are better. Bold and underline mark the best and second-best edited methods within each geometry; ties share a rank.}
\label{tab:app_toy_starplot_values}
\small
\setlength{\tabcolsep}{2.8pt}
\renewcommand{\arraystretch}{1.03}
\begin{tabular}{lrrrrrr}
\toprule
Method & Acc & F1 & $D_{\mathrm{toy}}$ & $U$ & Mean $C_1$ & Mean $C_2$ \\
\midrule
\multicolumn{7}{l}{\textit{Hypersphere}} \\
CMF & \underline{0.50000000} & 0.49944383 & 0.98280340 & 0.10500000 & \underline{0.02109016} & \textemdash \\
SFID & \underline{0.50000000} & 0.49776786 & 0.98110009 & 0.05250000 & 0.08129425 & \textemdash \\
SPD & \underline{0.50000000} & \underline{0.49986107} & \underline{0.98322745} & \underline{0.30500000} & 0.02415661 & \textemdash \\
IsoAct & \textbf{0.50833333} & \textbf{0.50802585} & \textbf{1.00000000} & \textbf{1.00000000} & \textbf{2.33146835e-17} & \textemdash \\
\midrule
\multicolumn{7}{l}{\textit{Lorentz}} \\
CMF & 1.00000000 & 1.00000000 & 0.00000000 & \underline{0.97750000} & \underline{3.66718367} & \textbf{0.00000000} \\
SFID & \textbf{0.50000000} & \textbf{0.49986107} & \textbf{1.00000000} & \textbf{1.00000000} & 15.20179613 & \textbf{0.00000000} \\
SPD & \underline{0.48333333} & \underline{0.47972028} & \underline{0.96319814} & \textbf{1.00000000} & 10.98409801 & \textbf{0.00000000} \\
IsoAct & \textbf{0.50000000} & \textbf{0.49986107} & \textbf{1.00000000} & \textbf{1.00000000} & \textbf{3.83915122e-15} & \textbf{0.00000000} \\
\midrule
\multicolumn{7}{l}{\textit{$\SE(3)$}} \\
CMF & 0.54166667 & 0.54138003 & 0.88688103 & 0.63000000 & 1.22492562 & 0.86775255 \\
SFID & \underline{0.45833333} & \underline{0.45829571} & \underline{0.95215898} & 0.58250000 & \underline{0.17973038} & \underline{0.00859062} \\
SPD & \textbf{0.48333333} & \textbf{0.48275862} & \textbf{1.00000000} & \underline{0.99250000} & 1.03584875 & 0.72990093 \\
IsoAct & \textbf{0.48333333} & \textbf{0.48275862} & \textbf{1.00000000} & \textbf{1.00000000} & \textbf{2.44253558e-16} & \textbf{1.28785871e-16} \\
\bottomrule
\end{tabular}
\end{table}

\begin{table}[tbp]
\centering
\caption{Additional controlled results. Nuisance columns show raw scores for Original/Oracle and absolute Oracle discrepancies for edited methods. Lower discrepancies, MAE, and residuals and higher $U$ are better. Bold/underline mark best/second-best edited methods within each geometry; ties share a rank at displayed precision. Table~\ref{tab:app_toy_starplot_values} gives higher-precision residuals.}
\label{tab:normalized_unit_norm_results_icml}
\small
\setlength{\tabcolsep}{2.5pt}
\renewcommand{\arraystretch}{1.04}
\begin{tabular}{lrrrrrrrr}
\toprule
 & \multicolumn{4}{c}{Nuisance metric / Oracle discrepancy} & \multicolumn{2}{c}{Task} & \multicolumn{2}{c}{Residual mean} \\
\cmidrule(lr){2-5}\cmidrule(lr){6-7}\cmidrule(lr){8-9}
Method & Acc & Recall & F1 & AUROC & MAE & $U$ & $C_1$ & $C_2$ \\
\midrule
\multicolumn{9}{l}{\textit{Hypersphere}} \\
\textcolor{gray}{Original} & \textcolor{gray}{1.0000} & \textcolor{gray}{1.0000} & \textcolor{gray}{1.0000} & \textcolor{gray}{1.0000} & \textcolor{gray}{0.0000} & \textcolor{gray}{1.0000} & \textcolor{gray}{0.0000} & \textcolor{gray}{\textemdash} \\
\textcolor{gray}{Oracle} & \textcolor{gray}{0.5083} & \textcolor{gray}{0.5083} & \textcolor{gray}{0.5080} & \textcolor{gray}{0.4956} & \textcolor{gray}{0.0000} & \textcolor{gray}{1.0000} & \textcolor{gray}{0.0000} & \textcolor{gray}{\textemdash} \\
CMF & \underline{0.0083} & \underline{0.0083} & 0.0086 & \underline{0.0023} & 3.2653 & 0.1050 & \underline{0.0211} & \textemdash \\
SFID & \underline{0.0083} & \underline{0.0083} & 0.0103 & \textbf{0.0000} & 13.1276 & 0.0525 & 0.0813 & \textemdash \\
SPD & \underline{0.0083} & \underline{0.0083} & \underline{0.0082} & 0.0109 & \underline{2.9891} & \underline{0.3050} & 0.0242 & \textemdash \\
IsoAct & \textbf{0.0000} & \textbf{0.0000} & \textbf{0.0000} & \textbf{0.0000} & \textbf{0.0000} & \textbf{1.0000} & \textbf{0.0000} & \textemdash \\
\midrule
\multicolumn{9}{l}{\textit{Lorentz}} \\
\textcolor{gray}{Original} & \textcolor{gray}{1.0000} & \textcolor{gray}{1.0000} & \textcolor{gray}{1.0000} & \textcolor{gray}{1.0000} & \textcolor{gray}{0.0000} & \textcolor{gray}{1.0000} & \textcolor{gray}{0.0000} & \textcolor{gray}{0.0000} \\
\textcolor{gray}{Oracle} & \textcolor{gray}{0.5000} & \textcolor{gray}{0.5000} & \textcolor{gray}{0.4999} & \textcolor{gray}{0.5008} & \textcolor{gray}{0.0000} & \textcolor{gray}{1.0000} & \textcolor{gray}{0.0000} & \textcolor{gray}{0.0000} \\
CMF & 0.5000 & 0.5000 & 0.5001 & 0.4992 & 0.0161 & \underline{0.9775} & \underline{3.6672} & \textbf{0.0000} \\
SFID & \textbf{0.0000} & \textbf{0.0000} & \textbf{0.0000} & \textbf{0.0000} & \textbf{0.0000} & \textbf{1.0000} & 15.2018 & \textbf{0.0000} \\
SPD & \underline{0.0167} & \underline{0.0167} & \underline{0.0201} & \underline{0.0125} & \underline{0.0009} & \textbf{1.0000} & 10.9841 & \textbf{0.0000} \\
IsoAct & \textbf{0.0000} & \textbf{0.0000} & \textbf{0.0000} & \textbf{0.0000} & \textbf{0.0000} & \textbf{1.0000} & \textbf{0.0000} & \textbf{0.0000} \\
\midrule
\multicolumn{9}{l}{\textit{$\SE(3)$}} \\
\textcolor{gray}{Original} & \textcolor{gray}{1.0000} & \textcolor{gray}{1.0000} & \textcolor{gray}{1.0000} & \textcolor{gray}{1.0000} & \textcolor{gray}{0.0000} & \textcolor{gray}{1.0000} & \textcolor{gray}{0.0000} & \textcolor{gray}{0.0000} \\
\textcolor{gray}{Oracle} & \textcolor{gray}{0.4833} & \textcolor{gray}{0.4833} & \textcolor{gray}{0.4828} & \textcolor{gray}{0.5133} & \textcolor{gray}{0.0000} & \textcolor{gray}{1.0000} & \textcolor{gray}{0.0000} & \textcolor{gray}{0.0000} \\
CMF & 0.0583 & 0.0584 & 0.0586 & 0.0067 & 0.4223 & 0.6300 & 1.2249 & 0.8678 \\
SFID & \underline{0.0250} & \underline{0.0250} & \underline{0.0245} & 0.0377 & 0.4636 & 0.5825 & \underline{0.1797} & \underline{0.0086} \\
SPD & \textbf{0.0000} & \textbf{0.0000} & \textbf{0.0000} & \underline{0.0009} & \underline{0.0104} & \underline{0.9925} & 1.0358 & 0.7299 \\
IsoAct & \textbf{0.0000} & \textbf{0.0000} & \textbf{0.0000} & \textbf{0.0000} & \textbf{0.0000} & \textbf{1.0000} & \textbf{0.0000} & \textbf{0.0000} \\
\bottomrule
\end{tabular}
\end{table}

\begin{figure}[t]
 \centering
 \includegraphics[width=0.79\linewidth]{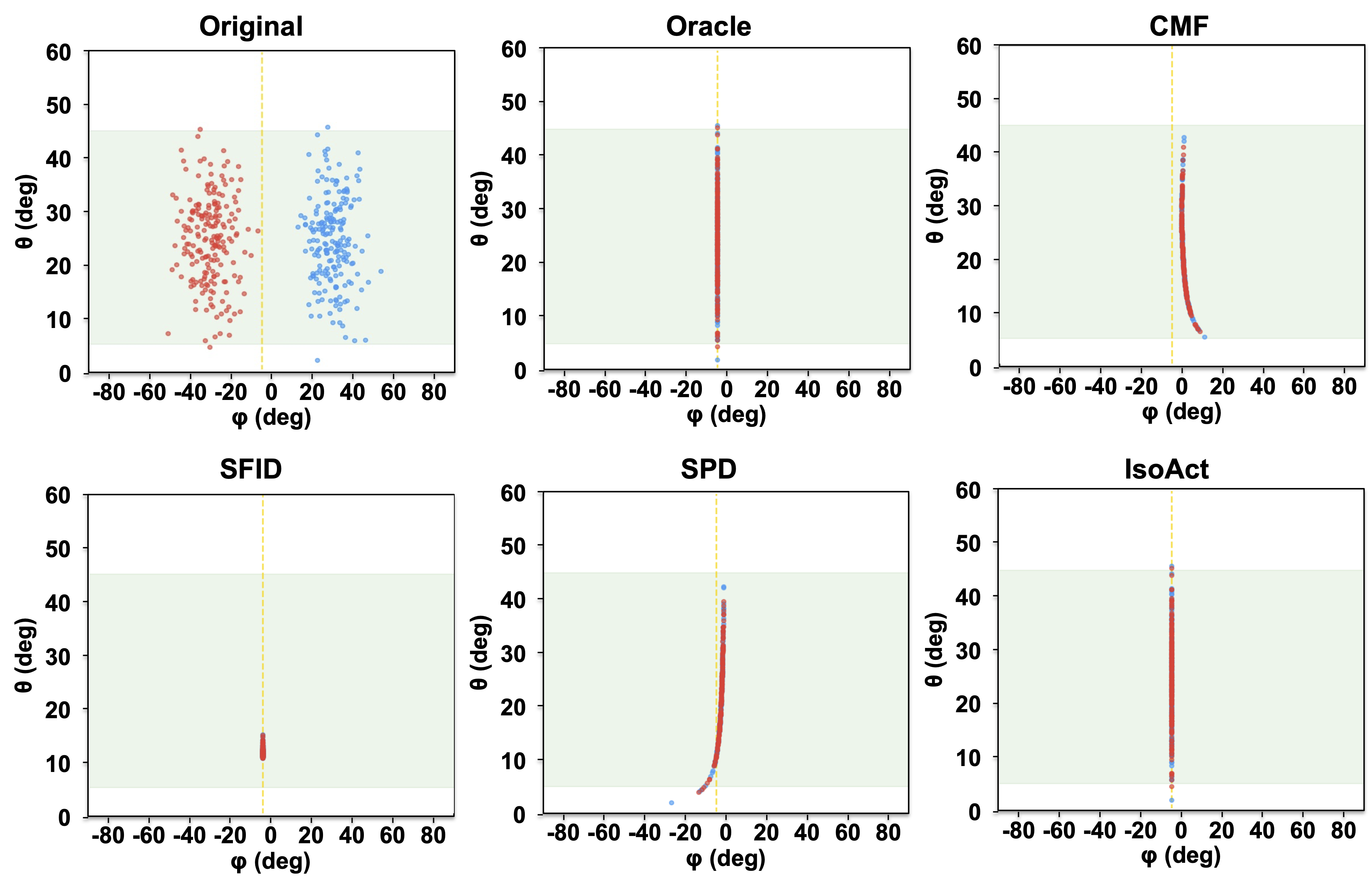}
 \caption{Hyperspherical mechanism check. The horizontal angle carries nuisance and the polar angle is the shared utility proxy. IsoAct collapses group separation while preserving the proxy. Shading is a visual guide; utility uses each sample's original-angle $2^\circ$ tolerance.}
 \label{fig:sphere_controlled_scatter}
\end{figure}

\begin{figure}[t]
 \centering
 \includegraphics[width=0.79\linewidth]{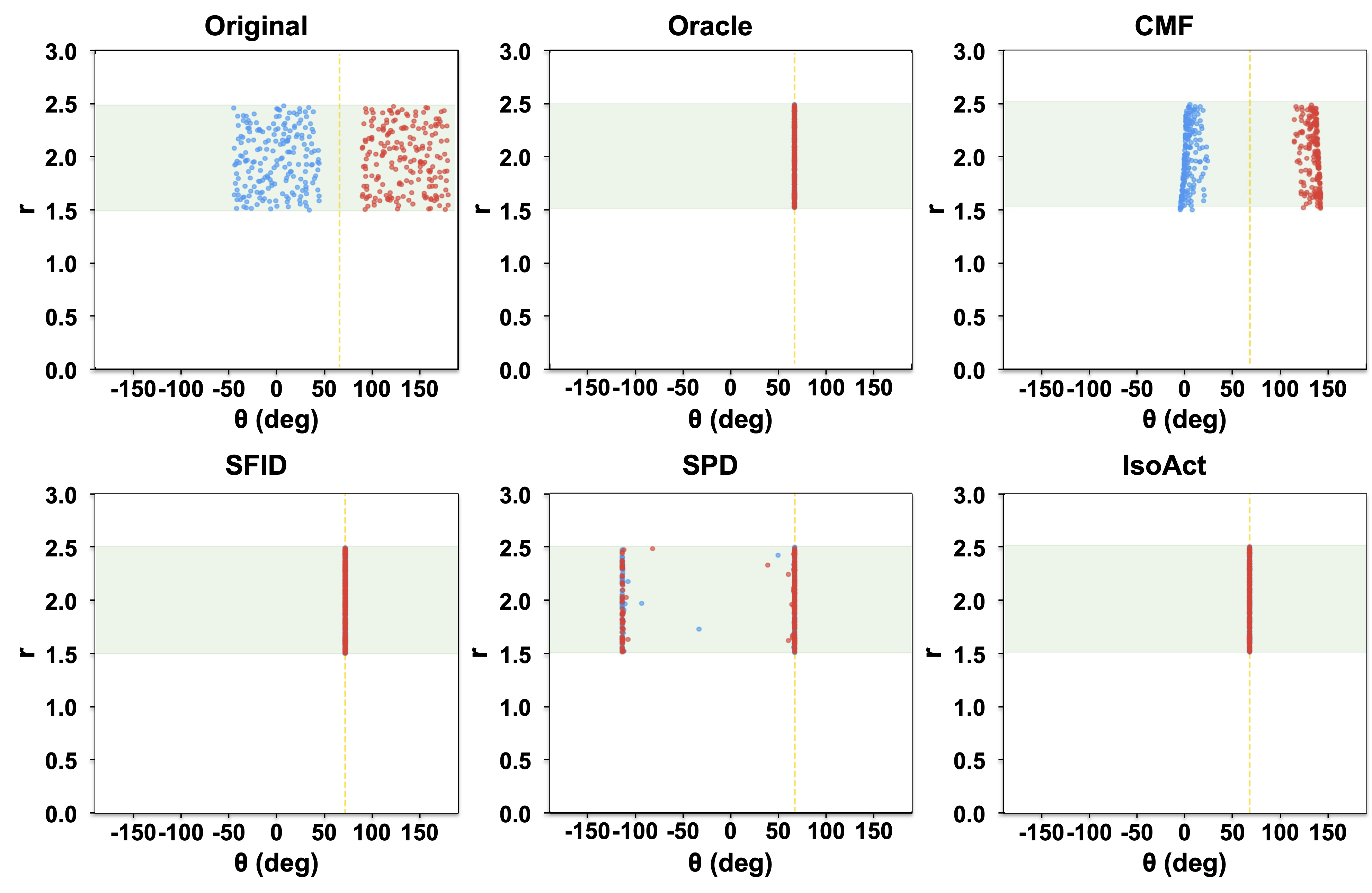}
 \caption{Lorentz mechanism check. IsoAct removes angular group separation while retaining the shared radial utility range. The dashed line marks the neutral angle and shading marks the shared range; the accompanying residuals assess manifold validity.}
 \label{fig:lorentz_controlled_scatter}
\end{figure}

Table~\ref{tab:app_toy_starplot_values} gives the probe scores and residuals underlying the starplots; Table~\ref{tab:normalized_unit_norm_results_icml} adds Oracle discrepancies and utility-proxy errors. IsoAct matches the Oracle probe metrics while satisfying the proxy criterion for every example. Figures~\ref{fig:sphere_controlled_scatter} and~\ref{fig:lorentz_controlled_scatter} complement the main $\SE(3)$ scatter with views of nuisance separation and shared utility structure.


\subsection{KLIFS ligand-pose evaluation}
\label{app:klifs_full_results_icml}

\paragraph{Data and tasks.}
The train, validation, and test splits contain 7,064, 2,189, and 2,596 poses, respectively, with seed 48 and disjoint kinase IDs. The nine nuisance groups are AGC, Atypical, CAMK, CK1, CMGC, Other, STE, TK, and TKL. Kinase group is a nuisance proxy for family-specific cues, while utility concerns conformation and ligand placement. Each input pose uses the packing in Section~\ref{app:se3_details_icml}.

The $\alpha$C-helix task encodes in/out states as 0/1 and excludes other states. Pocket metrics average three separately evaluated binary tasks: front, gate, and back occupancy. Structural-task F1 and Recall treat the minority class identified from training labels as positive; AUROC uses class-1 probabilities.

\paragraph{Predictors and selected settings.}
Nuisance prediction uses standardization and balanced logistic regression with at most 2,000 iterations. Each utility task uses standardization and a one-hidden-layer multilayer perceptron with 128 units, at most 300 iterations, and seed 42. Predictors are refitted on each transformed training set and evaluated on the associated validation and test sets.

The selected IsoAct configuration uses translation-only actions with $K=5$ and $\alpha=2$. At each direction, the full packed-pose discrepancy determines the translation update. The selected CMF adaptation has latent dimension 8, geometry weight 0.1, and reconstruction weight 5. SFID and SPD use strengths 2 and 0.75, respectively; SPD requests ten directions.

Table~\ref{tab:real_world_full_icml} reports the selected configurations' test nuisance Accuracy and macro-F1, structural-task Recall and AUROC, and mean rotation-constraint residuals. The shuffled-label Null supplies the nuisance-predictability target. Section~\ref{app:constraint_diagnostics} defines the residuals.

\subsection{\texorpdfstring{$\SE(3)$}{SE(3)} Pareto display}
\label{app:pareto_details}

Each point in Figure~\ref{fig:se3_pareto} reports the test performance of one configuration. With nuisance metrics $M_m$, Original values $O_m$, and Null values $N_m$, the horizontal coordinate is
\[
 D_{\mathrm{Pareto}}=\frac12\sum_{m\in\{\mathrm{Acc},\mathrm{macro\mbox{-}F1}\}}
 \left(1-\frac{|M_m-N_m|}{|O_m-N_m|}\right).
\]
The vertical coordinate averages four utility-retention ratios,
\[
 U_{\mathrm{Pareto}}=\frac14\sum_{t\in\{\alpha C,\mathrm{Pocket}\}}
 \left(\frac{\mathrm{Recall}_t}{\mathrm{Recall}^{\mathrm{Original}}_t}
 +\frac{\mathrm{AUROC}_t}{\mathrm{AUROC}^{\mathrm{Original}}_t}\right).
\]
Pocket component metrics are averaged before forming the ratios. The plot uses linear axes over $0\le D\le0.8$ and $0.8\le U\le1.1$.

The pools vary IsoAct's action family, retained rank, and strength; SPD's rank and strength; SFID's strength; and CMF's latent dimension and geometry/reconstruction weights. Edit-strength ranges overlap where applicable. The dashed line connects the highlighted IsoAct configurations to summarize their observed trade-off.

\subsection{CLIP, MERU, FairFace, and Flickr30K}
\label{app:face_details}

\paragraph{Representations and subsets.}
CLIP ViT-B/32 supplies 512-dimensional unit-normalized image vectors. MERU ViT-B/16, with input resolution 224, supplies 512 spatial coordinates on the Lorentz model with curvature $-c\approx-0.1$.

For each protected attribute, fixed FairFace subsets contain 3,500 training, 1,500 validation, and 1,500 test examples from 86,744 embeddings per model. Sampling without replacement approximately preserves class proportions using rounded class allocations. The respective subset seeds are 42, 43, and 44. Each attribute uses identical subsets for CLIP and MERU.

Age has nine classes: 0 to 2, 3 to 9, 10 to 19, 20 to 29, 30 to 39, 40 to 49, 50 to 59, 60 to 69, and more than 70. Gender labels are Female and Male. Race labels are East Asian, Indian, Black, White, Middle Eastern, Latino/Hispanic, and Southeast Asian. Each set of dataset annotations defines a separate nuisance-prediction task.

\paragraph{Nuisance probes and Null.}
Each transformed training set is used to fit standardization and an L2-regularized logistic-regression probe with \texttt{lbfgs}, balanced class weights, inverse regularization strength 1, at most 2,000 iterations, and seed 42. The same fitted probe supplies validation and test Accuracy and macro-F1. Null is estimated by averaging scores over deterministic label permutations against the Original predictions.

\paragraph{Retrieval utility.}
Flickr30K's Karpathy validation split contains 1,014 images and 5,070 text queries; its test split contains 1,000 images and 5,000 queries. The learned transformation edits image representations, while text representations remain fixed. Each query retrieves its corresponding image from the full gallery. CLIP uses dot-product similarity. For MERU, scores take the form $\operatorname{softmax}(100VT^\top)$, with image vectors as rows of $V$ and text vectors as rows of $T$.

R@$k$ measures top-$k$ retrieval success. Exact score ties use an expected-uniform policy: if $g$ images score above the target and $m$ images, including the target, tie with it, the query contributes $\operatorname{clip}(k-g,0,m)/m$. Averaging over all queries gives R@1, R@5, and R@10.

\paragraph{Reported coordinates.}
For edited, Original, and Null nuisance metrics $M_m,O_m,N_m$, define $D_m=1-|M_m-N_m|/|O_m-N_m|$. Debias averages $D_{\mathrm{Acc}}$ and $D_{\mathrm{macro\mbox{-}F1}}$. Retrieval retention is
\[
 U_{\mathrm{ret}}=\frac13\sum_{k\in\{1,5,10\}}
 \frac{\mathrm{R@}k}{\mathrm{R@}k_{\mathrm{Original}}}.
\]
Ratios are formed separately before averaging, using the corresponding Original and Null references and finite, nonzero denominators. Negative Debias values indicate greater normalized distance from Null than Original; retention above one indicates a gain relative to Original.

\paragraph{Selected configurations.}
Table~\ref{tab:app_face_configs} lists the selected IsoAct rank, ridge, strengths, and composition order. All use forward direction order. CLIP race repeats its six-direction sequence twice; MERU composes angular and translation actions within each direction.

\begin{table}[tbp]
\centering
\caption{Selected IsoAct settings for CLIP and MERU. $K$ is the retained rank and $\rho$ the FDA ridge. Ang. and Trans. denote angular and translation actions composed within each direction; a dash indicates a single spherical action.}
\label{tab:app_face_configs}
\small
\setlength{\tabcolsep}{3.5pt}
\begin{tabular}{llrrrll}
\toprule
Model & Attribute & $K$ & $\rho$ & Passes & Strengths & \makecell{Within-axis\\order} \\
\midrule
CLIP & Age    & 7 & 0.1  & 1 & $\alpha=1$ & \textemdash \\
     & Gender & 1 & 0.01 & 1 & $\alpha=1$ & \textemdash \\
     & Race   & 6 & 0.1  & 2 & $\alpha=1$ & \textemdash \\
\midrule
MERU & Age    & 8 & 0.01 & 1 & $\alpha_{\mathrm{ang}}=0.75$, $\alpha_{\mathrm{tr}}=1$ & Ang. $\to$ Trans. \\
     & Gender & 1 & 0.03 & 1 & $\alpha_{\mathrm{ang}}=0.5$, $\alpha_{\mathrm{tr}}=1$ & Trans. $\to$ Ang. \\
     & Race   & 6 & 0.1  & 1 & $\alpha_{\mathrm{ang}}=0.5$, $\alpha_{\mathrm{tr}}=1$ & Trans. $\to$ Ang. \\
\bottomrule
\end{tabular}
\end{table}

\begin{table}[tbp]
\centering
\caption{CLIP raw test probe and retrieval metrics. F1 denotes macro-F1. Bold/underline mark best/second-best methods within each attribute: smaller absolute distance to Null for Acc/F1, higher values for retrieval. Original and Null are unranked; ties share a rank. Table~\ref{tab:clip_meru_main} reports normalized summaries.}
\label{tab:app_clip_metrics}
\small
\setlength{\tabcolsep}{5pt}
\renewcommand{\arraystretch}{1.0}
\begin{tabular}{lrrrrr}
\toprule
Method & Acc & F1 & R@1 & R@5 & R@10 \\
\midrule
\multicolumn{6}{l}{\textit{Age}} \\
\textcolor{gray}{Original} & \textcolor{gray}{0.2987} & \textcolor{gray}{0.2614} & \textcolor{gray}{0.5876} & \textcolor{gray}{0.8350} & \textcolor{gray}{0.9004} \\
\textcolor{gray}{Null} & \textcolor{gray}{0.1477} & \textcolor{gray}{0.1060} & \textcolor{gray}{\textemdash} & \textcolor{gray}{\textemdash} & \textcolor{gray}{\textemdash} \\
CMF & 0.1920 & 0.1742 & 0.0012 & 0.0070 & 0.0166 \\
SFID & 0.2967 & 0.2599 & \textbf{0.5892} & \textbf{0.8354} & \textbf{0.8990} \\
SPD & \underline{0.1860} & \underline{0.1431} & 0.5764 & 0.8268 & 0.8906 \\
IsoAct & \textbf{0.1320} & \textbf{0.1137} & \underline{0.5770} & \underline{0.8290} & \underline{0.8934} \\
\midrule
\multicolumn{6}{l}{\textit{Gender}} \\
\textcolor{gray}{Original} & \textcolor{gray}{0.7720} & \textcolor{gray}{0.7714} & \textcolor{gray}{0.5876} & \textcolor{gray}{0.8350} & \textcolor{gray}{0.9004} \\
\textcolor{gray}{Null} & \textcolor{gray}{0.5008} & \textcolor{gray}{0.4995} & \textcolor{gray}{\textemdash} & \textcolor{gray}{\textemdash} & \textcolor{gray}{\textemdash} \\
CMF & \underline{0.6507} & \underline{0.6494} & 0.0028 & 0.0102 & 0.0198 \\
SFID & 0.7613 & 0.7610 & 0.5706 & 0.8198 & 0.8856 \\
SPD & \textbf{0.5480} & \textbf{0.5478} & \underline{0.5810} & \underline{0.8322} & \underline{0.8968} \\
IsoAct & 0.6567 & 0.6558 & \textbf{0.5874} & \textbf{0.8350} & \textbf{0.9016} \\
\midrule
\multicolumn{6}{l}{\textit{Race}} \\
\textcolor{gray}{Original} & \textcolor{gray}{0.3973} & \textcolor{gray}{0.3910} & \textcolor{gray}{0.5876} & \textcolor{gray}{0.8350} & \textcolor{gray}{0.9004} \\
\textcolor{gray}{Null} & \textcolor{gray}{0.1457} & \textcolor{gray}{0.1430} & \textcolor{gray}{\textemdash} & \textcolor{gray}{\textemdash} & \textcolor{gray}{\textemdash} \\
CMF & 0.2327 & 0.2116 & 0.0022 & 0.0098 & 0.0170 \\
SFID & 0.3993 & 0.3932 & \textbf{0.5840} & \textbf{0.8354} & \textbf{0.8986} \\
SPD & \textbf{0.1607} & \textbf{0.1454} & 0.5800 & 0.8274 & 0.8908 \\
IsoAct & \underline{0.2033} & \underline{0.1844} & \underline{0.5818} & \underline{0.8310} & \underline{0.8952} \\
\bottomrule
\end{tabular}
\end{table}

\begin{table}[tbp]
\centering
\caption{MERU raw test probe and retrieval metrics. F1 denotes macro-F1. Bold/underline mark best/second-best methods within each attribute: smaller absolute distance to Null for Acc/F1, higher values for retrieval. Original and Null are unranked; ties share a rank. Table~\ref{tab:clip_meru_main} reports normalized summaries.}
\label{tab:app_meru_metrics}
\small
\setlength{\tabcolsep}{5pt}
\renewcommand{\arraystretch}{1.0}
\begin{tabular}{lrrrrr}
\toprule
Method & Acc & F1 & R@1 & R@5 & R@10 \\
\midrule
\multicolumn{6}{l}{\textit{Age}} \\
\textcolor{gray}{Original} & \textcolor{gray}{0.2747} & \textcolor{gray}{0.2465} & \textcolor{gray}{0.2096} & \textcolor{gray}{0.4242} & \textcolor{gray}{0.5216} \\
\textcolor{gray}{Null} & \textcolor{gray}{0.1439} & \textcolor{gray}{0.1055} & \textcolor{gray}{\textemdash} & \textcolor{gray}{\textemdash} & \textcolor{gray}{\textemdash} \\
CMF & 0.2573 & 0.2359 & 0.0372 & 0.1188 & 0.1778 \\
SFID & 0.2587 & 0.2424 & \textbf{0.2030} & \textbf{0.4194} & \textbf{0.5160} \\
SPD & \underline{0.1373} & \underline{0.1236} & \underline{0.1986} & \underline{0.4146} & \underline{0.5094} \\
IsoAct & \textbf{0.1413} & \textbf{0.1199} & 0.1900 & 0.4114 & 0.5082 \\
\midrule
\multicolumn{6}{l}{\textit{Gender}} \\
\textcolor{gray}{Original} & \textcolor{gray}{0.7440} & \textcolor{gray}{0.7435} & \textcolor{gray}{0.2096} & \textcolor{gray}{0.4242} & \textcolor{gray}{0.5216} \\
\textcolor{gray}{Null} & \textcolor{gray}{0.4976} & \textcolor{gray}{0.4966} & \textcolor{gray}{\textemdash} & \textcolor{gray}{\textemdash} & \textcolor{gray}{\textemdash} \\
CMF & 0.6827 & 0.6818 & 0.0428 & 0.1270 & 0.1968 \\
SFID & 0.7440 & 0.7435 & \textbf{0.2070} & \textbf{0.4244} & \underline{0.5190} \\
SPD & \underline{0.6000} & \underline{0.5996} & \underline{0.2048} & \underline{0.4180} & \textbf{0.5206} \\
IsoAct & \textbf{0.5247} & \textbf{0.5242} & 0.1992 & 0.4108 & 0.5158 \\
\midrule
\multicolumn{6}{l}{\textit{Race}} \\
\textcolor{gray}{Original} & \textcolor{gray}{0.3987} & \textcolor{gray}{0.3900} & \textcolor{gray}{0.2096} & \textcolor{gray}{0.4242} & \textcolor{gray}{0.5216} \\
\textcolor{gray}{Null} & \textcolor{gray}{0.1443} & \textcolor{gray}{0.1421} & \textcolor{gray}{\textemdash} & \textcolor{gray}{\textemdash} & \textcolor{gray}{\textemdash} \\
CMF & 0.1967 & 0.1902 & 0.0024 & 0.0088 & 0.0152 \\
SFID & 0.3860 & 0.3793 & \textbf{0.1994} & \textbf{0.4082} & \textbf{0.5076} \\
SPD & \underline{0.1860} & \underline{0.1735} & 0.1892 & \underline{0.3978} & 0.4986 \\
IsoAct & \textbf{0.1507} & \textbf{0.1445} & \underline{0.1970} & 0.3940 & \underline{0.5038} \\
\bottomrule
\end{tabular}
\end{table}

Tables~\ref{tab:app_clip_metrics} and~\ref{tab:app_meru_metrics} provide the raw test metrics underlying the main table's normalized summaries.

\subsection{Manifold-constraint diagnostics}
\label{app:constraint_diagnostics}

For a unit-sphere output, $C_1(x)=|\|x\|_2-1|$. On the Lorentz model with curvature $-c$, where $c>0$, a point $x=(x_0,v)$ has residuals
\[
 C_1(x)=|-x_0^2+\|v\|_2^2+c^{-1}|,
 \qquad C_2(x)=\max(0,-x_0).
\]
For a pose with rotation block $R$,
\[
 C_1(R)=\|R^\top R-I_3\|_F,
 \qquad C_2(R)=|\det R-1|.
\]
We report sample means $\overline C_1,\overline C_2$. Real-world diagnostics use the selected FairFace test embeddings (1,500 per attribute) and KLIFS test poses (2,596). The Lorentz terms measure quadratic-constraint error and negative-time magnitude; the rotation terms measure orthogonality and determinant error.

\begin{table}[tbp]
\centering
\caption{Mean manifold-constraint residuals on selected test representations. IsoAct maintains unit-norm and rotation constraints to numerical precision. Lower is better. Bold/underline mark best/second-best methods within each setting, excluding Original and all MERU residuals. MERU's Lorentz quadratic and upper-sheet residuals are reported separately.}
\label{tab:app_constraints}
\small
\setlength{\tabcolsep}{5pt}
\renewcommand{\arraystretch}{1.0}
\begin{tabular}{llrr}
\toprule
Setting & Method & Mean $C_1$ & Mean $C_2$ \\
\midrule
\textcolor{gray}{KLIFS} & \textcolor{gray}{Original} & \textcolor{gray}{6.5065e-08} & \textcolor{gray}{2.2077e-08} \\
 & CMF & 3.6900e-01 & 1.1368e-01 \\
 & SFID & \underline{6.5065e-08} & \underline{2.2077e-08} \\
 & SPD & 9.4948e-01 & 3.7450e-01 \\
 & IsoAct & \textbf{6.9510e-16} & \textbf{3.5723e-16} \\
\midrule
\textcolor{gray}{CLIP age} & \textcolor{gray}{Original} & \textcolor{gray}{2.2612e-08} & \textcolor{gray}{\textemdash} \\
 & CMF & 8.8207e-02 & \textemdash \\
 & SFID & \underline{3.8070e-03} & \textemdash \\
 & SPD & 1.8327e-02 & \textemdash \\
 & IsoAct & \textbf{6.3230e-08} & \textemdash \\
\midrule
\textcolor{gray}{CLIP gender} & \textcolor{gray}{Original} & \textcolor{gray}{2.2935e-08} & \textcolor{gray}{\textemdash} \\
 & CMF & 7.0113e-02 & \textemdash \\
 & SFID & \underline{8.1287e-03} & \textemdash \\
 & SPD & 1.0768e-02 & \textemdash \\
 & IsoAct & \textbf{3.5303e-08} & \textemdash \\
\midrule
\textcolor{gray}{CLIP race} & \textcolor{gray}{Original} & \textcolor{gray}{2.2181e-08} & \textcolor{gray}{\textemdash} \\
 & CMF & 1.1123e-01 & \textemdash \\
 & SFID & \underline{8.3426e-03} & \textemdash \\
 & SPD & 5.3183e-02 & \textemdash \\
 & IsoAct & \textbf{6.1190e-08} & \textemdash \\
\midrule
\textcolor{gray}{MERU age} & \textcolor{gray}{Original} & \textcolor{gray}{7.9818e-16} & \textcolor{gray}{0.0000e+00} \\
 & CMF & 7.6620e-16 & 0.0000e+00 \\
 & SFID & 7.8515e-16 & 0.0000e+00 \\
 & SPD & 7.8397e-16 & 0.0000e+00 \\
 & IsoAct & 7.7804e-16 & 0.0000e+00 \\
\midrule
\textcolor{gray}{MERU gender} & \textcolor{gray}{Original} & \textcolor{gray}{8.1239e-16} & \textcolor{gray}{0.0000e+00} \\
 & CMF & 7.8278e-16 & 0.0000e+00 \\
 & SFID & 7.7804e-16 & 0.0000e+00 \\
 & SPD & 7.6383e-16 & 0.0000e+00 \\
 & IsoAct & 7.6383e-16 & 0.0000e+00 \\
\midrule
\textcolor{gray}{MERU race} & \textcolor{gray}{Original} & \textcolor{gray}{7.8870e-16} & \textcolor{gray}{0.0000e+00} \\
 & CMF & 8.0647e-16 & 0.0000e+00 \\
 & SFID & 7.6383e-16 & 0.0000e+00 \\
 & SPD & 7.8989e-16 & 0.0000e+00 \\
 & IsoAct & 7.8515e-16 & 0.0000e+00 \\
\bottomrule
\end{tabular}
\end{table}

MERU represents each spatial vector $v$ with dependent time coordinate $x_0=\sqrt{\|v\|^2+c^{-1}}>0$. This satisfies the hyperboloid constraint in exact arithmetic, explaining the numerical-precision $C_1$ and zero $C_2$ across methods. The main table averages $\overline C_1$ and $\overline C_2$, while Table~\ref{tab:app_constraints} lists both residuals separately.

\section{Compute Resources}
\label{app:compute_resources_icml}

Experiments on frozen representations used an Ubuntu CPU node with an Intel Xeon Gold 6326 processor at 2.90 GHz, 64 CPU threads, and 251 GiB RAM. The final evaluation environment included Python 3.10.19, NumPy 1.26.4, SciPy 1.15.3, and scikit-learn 1.7.2. Dataset splits, evaluation predictors, and selected settings are specified in Appendix~\ref{app:experiments}.



\end{document}